\documentclass[lettersize,journal]{IEEEtran}
\usepackage{amsmath,amsfonts}
\usepackage{algorithmic}
\usepackage{algorithm}
\usepackage{array}
\usepackage{textcomp}
\usepackage{stfloats}
\usepackage{url}
\usepackage{verbatim}
\usepackage{graphicx}
\usepackage{cite}
\usepackage{multirow}
\usepackage{subcaption}
\usepackage{enumitem}
\usepackage{booktabs}
\usepackage{alphalph}
\usepackage[flushleft]{threeparttable}
\usepackage{pifont}
\usepackage{cite}
\usepackage{amsmath,amssymb,amsfonts}
\usepackage{graphicx}
\usepackage{textcomp}
\usepackage{xcolor}

\begin{document}

\title{Unraveling Spatio-Temporal Foundation Models via the Pipeline Lens: A Comprehensive Review}

\author{Yuchen~Fang, Hao~Miao, Yuxuan~Liang, Liwei~Deng, Yue~Cui, Ximu~Zeng, Yuyang~Xia, Yan~Zhao$^*$,~\IEEEmembership{Senior~Member,~IEEE}, Torben~Bach Pedersen,~\IEEEmembership{Senior~Member,~IEEE}, Christian~S.
Jensen,~\IEEEmembership{Fellow,~IEEE}, Xiaofang~Zhou,~\IEEEmembership{Fellow,~IEEE}, and~Kai~Zheng$^*$,~\IEEEmembership{Senior~Member,~IEEE}.
\thanks{Manuscript received xxx; revised xxx. This work is partially supported by NSFC (No. 62472068), Shenzhen Municipal Science and Technology R\&D Funding Basic Research Program (JCYJ20210324133607021), and Municipal Government of Quzhou under Grant (2023D044), and Key Laboratory of Data Intelligence and Cognitive Computing, Longhua District, Shenzhen. \textit{(Corresponding authors: Yan Zhao; Kai Zheng.)}}
\thanks{\IEEEcompsocthanksitem Yuchen Fang, Liwei Deng, Ximu Zeng, Yuyang Xia, Yan Zhao, and Kai Zheng are with the University of Electronic Science and Technology of China, Chengdu, China (e-mail: fangyuchen@std.uestc.edu.cn; deng\_liwei@std.uestc.edu.cn; ximuzeng@std.uestc.edu.cn; xiayuyang@std.uestc.edu.cn; zhaoyan@uestc.edu.cn; zhengkai@uestc.edu.cn).
\IEEEcompsocthanksitem Hao Miao, Torben Bach Pedersen, and Christian S. Jensen are with the Aalborg University, Aalborg, Denmark (e-mail: haom@cs.aau.dk; tbp@cs.aau.dk; csj@cs.aau.dk).
\IEEEcompsocthanksitem Yuxuan Liang is with the Hong Kong University of Science and Technology (Guangzhou), Guangzhou, China (e-mail: yuxliang@outlook.com).
\IEEEcompsocthanksitem Yue Cui and Xiaofang Zhou are with the Hong Kong University of Science and Technology, Hong Kong, China (e-mail: ycuias@cse.ust.hk; zxf@cse.ust.hk).
 }
}

\markboth{Journal of \LaTeX\ Class Files,~Vol.~14, No.~8, August~2021}%
{Shell \MakeLowercase{\textit{et al.}}: A Sample Article Using IEEEtran.cls for IEEE Journals}

\IEEEpubid{0000--0000/00\$00.00~\copyright~2021 IEEE}

\maketitle

\begin{abstract}
Spatio-temporal data proliferates in numerous real-world domains, such as transportation, weather, and energy. Spatio-temporal deep learning models aims to utilize useful patterns in such data to support tasks like prediction, imputation, and anomaly detection. However, previous \emph{one-to-one} deep learning models designed for specific tasks typically require separate training for each use case, leading to increased computational and storage costs. To address this issue, \emph{one-to-many} spatio-temporal foundation models have emerged, offering a unified framework capable of solving multiple spatio-temporal tasks. These foundation models achieve remarkable success by learning general knowledge with spatio-temporal data or transferring the general capabilities of pre-trained language models. While previous surveys have explored spatio-temporal data and methodologies separately, they have ignored a comprehensive examination of how foundation models are designed, selected, pre-trained, and adapted. As a result, the overall pipeline for spatio-temporal foundation models remains unclear. To bridge this gap, we innovatively provide an up-to-date review of previous spatio-temporal foundation models from the pipeline perspective. The pipeline begins with an introduction to different types of spatio-temporal data, followed by details of data preprocessing and embedding techniques. The pipeline then presents a novel data property taxonomy to divide existing methods according to data sources and dependencies, providing efficient and effective model design and selection for researchers. On this basis, we further illustrate the training objectives of primitive models, as well as the adaptation techniques of transferred models. Overall, our survey provides a clear and structured pipeline to understand the connection between core elements of spatio-temporal foundation models while guiding researchers to get started quickly. Additionally, we introduce emerging opportunities such as multi-objective training in the field of spatio-temporal foundation models, providing valuable insights for researchers and practitioners.

\textit{GitHub Repository: https://github.com/LMissher/Awesome-Spatio-Temporal-Foundation-Models}
\end{abstract}

\begin{IEEEkeywords}
foundation models, spatio-temporal data, pre-training, adaption.
\end{IEEEkeywords}

\begin{figure}[ht]
  \centering
\includegraphics[width=\linewidth]{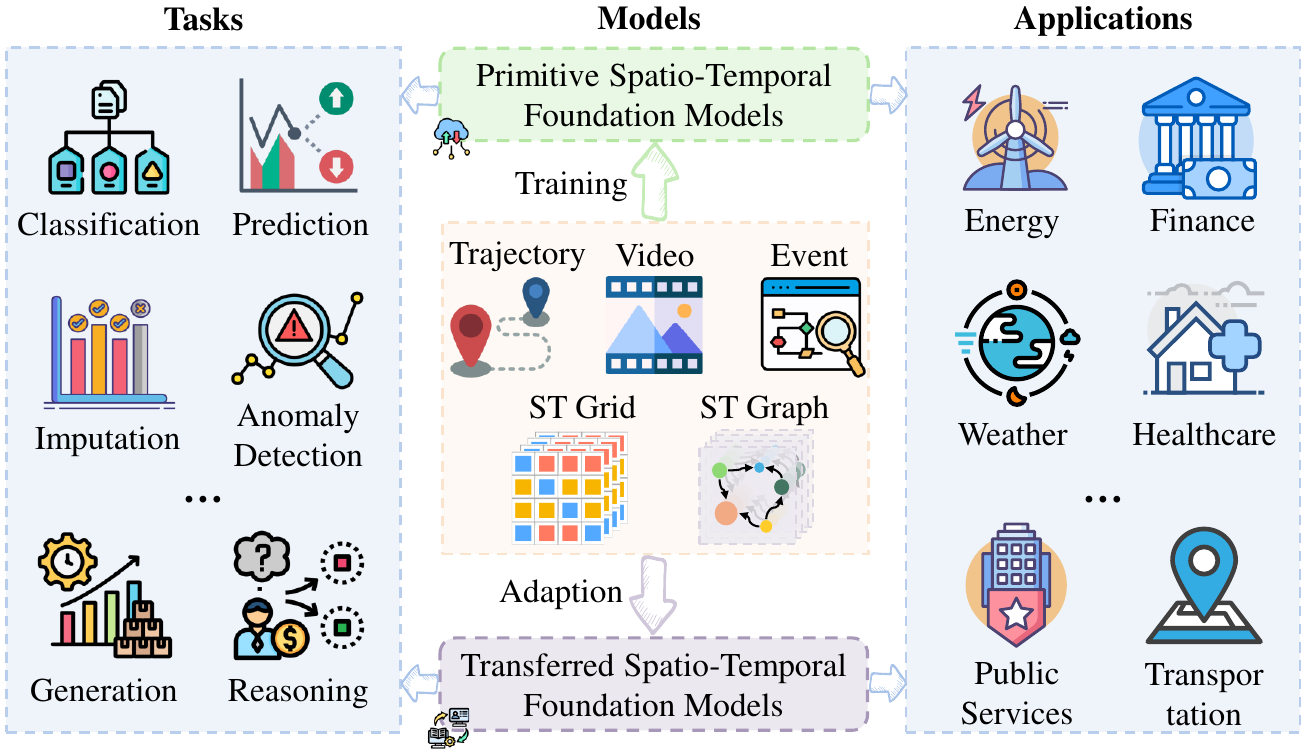}
  \caption{The paradigm of spatio-temporal foundation models (STFMs).}
  \label{param}
\end{figure}
\section{Introduction}
Spatio-temporal data are continuously generated from various real-world domains including transportation, energy, and weather. These data inherently exhibit an intricate temporal evolution over time and complex spatial interactions across different regions~\cite{jin2023spatio}. Various types of spatio-temporal data (\emph{e.g.}, trajectory data, traffic data, and video data) share common challenges in capturing spatio-temporal dependencies, requiring specialized methods to competently extract inherent correlations. The mining and analysis of these spatio-temporal correlations play a crucial role in building intelligent systems, enabling real-world applications to support decision-making in fundamental tasks including planning, reasoning, and anomaly detection.
\IEEEpubidadjcol

In recent years, considerable progress has been made in spatio-temporal data mining by \emph{one-to-one} specialized models based on the development of deep learning. These methods lie in the spatio-temporal modeling capabilities of sequential and spatial neural networks, such as recurrent neural networks (RNNs)~\cite{lv2018lc}, Transformers~\cite{fang2022learning}, convolution neural networks (CNNs)~\cite{he2017mask}, and graph neural networks (GNNs)~\cite{jiang2023uncertainty}. Nevertheless, addressing the wide range of spatio-temporal tasks across diverse applications requires training numerous task-specific models, which demands substantial computational resources and incurs significant costs. Fortunately, with the advent of self-supervised learning strategies and the discovery of scaling laws~\cite{kaplan2020scaling}, foundation models have been designed in the natural language processing and computer vision communities to universally solve multiple tasks through resource-efficient few-shot fine-tuning or even resource-free zero-shot prompting without additional training~\cite{hulora,wei2022chain}.

\begin{table*}[t]
    \aboverulesep=0ex
    \belowrulesep=0ex
    \caption{Comparison between our survey and related surveys. The spatio-temporal data types include trajectories (T), events (E), ST grids (G), videos (V), ST graphs (H). For training primitive models, objectives include regression (R), masked modeling (M), contrastive (C), and diffusion (D). Moreover, the adaption techniques for transferred models include prompt engineering (P), feature enhancement (F), cross-domain alignment (A), and supervised fine-tuning (S).}
    \centering    
    \resizebox{1.0\linewidth}{!}{
    \begin{tabular}{c|cc|ccc|c}
    \toprule
        Survey & Data Types & Data Harmonization & Model Taxonomy & Training Objectives & Transfer Adaption & Pipeline\\
    \midrule
        Jin et al.~\cite{jin2023large} & E,G,V & \ding{56} & Data Type & M,C,D & P,A,S & \ding{56}\\
        Liang et al.~\cite{liang2024foundation} & T,E,G,H & \ding{56} & Methodology & R,M,C,D & P,A,S & \ding{56}\\
        Zhang et al.~\cite{zhang2024urban} & T,E,G,H & \ding{56} & Data Type & R,M,C & P,A,S & \ding{56}\\
        Goodge et al.~\cite{goodge2025spatio} & T,E,G,H & \ding{56} & Methodology & R,M & A,S & \ding{56}\\
        Liang et al.~\cite{liang2025foundation} & T,E,G,H & \ding{56} & Methodology & M,C,D & P,F,A,S & \ding{56}\\
    \midrule
        Ours & T,E,G,V,H & \ding{52} & Data Property & R,M,C,D & P,F,A,S & \ding{52}\\
    \bottomrule
    \end{tabular}
    }
    \label{intro:comp}
\end{table*}
With the remarkable success of foundation models in natural language processing (\emph{e.g.}, ChatGPT), the concept of \emph{one-to-many} foundation models has been introduced as an attractive and promising direction in spatio-temporal communities. As illustrated in Fig.~\ref{param}, the goal of spatio-temporal foundation models is to learn general spatio-temporal knowledge within a single universal model. This enables the same model to handle a wide range of spatio-temporal tasks across different applications and objectives, significantly reducing the reliance on numerous task-specific models and thus lowering both training and storage costs. By increasing the training scale of spatio-temporal data and using general self-supervised learning objectives to derive \emph{primitive} foundation models, or transferring the general knowledge of pre-trained foundation models from other fields such as natural language processing to derive \emph{transferred} foundation models, the effectiveness of current spatio-temporal foundation models has been validated on various tasks, showing the promising prospect of the universal framework to advance this field.

Despite recent improvements of spatio-temporal foundation models (STFMs), existing surveys on this topic still face several key challenges. 1) \textbf{Weak linkage between data and models}: As illustrated in Table~\ref{intro:comp}, while previous surveys do describe various types of spatio-temporal data, they often overlook critical steps in the data harmonization process such as embedding techniques. This omission creates confusion regarding how spatio-temporal data is effectively aligned with foundation models. 2) \textbf{Lack of property consideration}: Prior surveys tend to adopt coarse-grained classifications of STFMs (\emph{e.g.}, data type and deep learning methodology perspectives) without explaining why similar methodologies are applied to different data types that share common characteristics. These taxonomies ignore deep insights from data properties in selecting or designing foundation models. 3) \textbf{Fragmented presentation}: Existing surveys tend to discuss spatio-temporal data, foundation models, training objectives, and transfer adaptation techniques in isolation. This siloed approach prevents a cohesive understanding of what models, objectives, and adaptation strategies should be utilized for different spatio-temporal tasks, datasets, and real-world applications.

To address the issue of fragmented descriptions, our survey offers a comprehensive examination of the entire pipeline of spatio-temporal foundation models (STFMs), systematically presenting the workflow from data harmonization and model conception to training, adaptation, and real-world application.

\begin{figure*}[t]
  \centering
\includegraphics[width=0.8\linewidth]{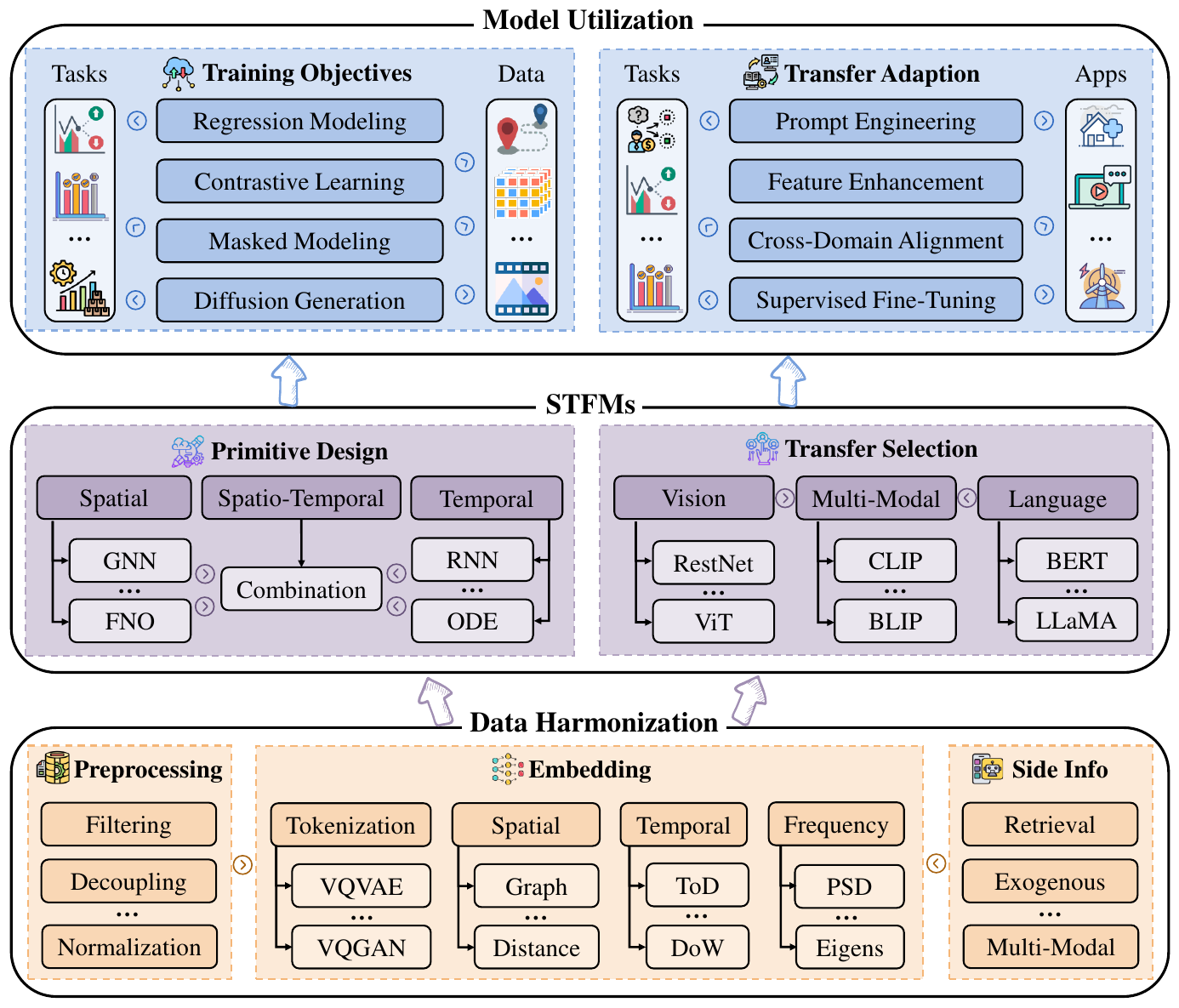}
  \caption{The pipeline of STFMs.}
  \label{pipeline_taxonomy}
\end{figure*}

In addition to a brief overview of spatio-temporal data and available datasets, our survey—illustrated in the bottom of Fig.~\ref{pipeline_taxonomy}—provides a detailed account of data preprocessing, embedding techniques, and side information associated with various spatio-temporal data types, thereby completing the first stage of the STFM pipeline: \textbf{data harmonization}. By leveraging side information and appropriate preprocessing methods, the quality of spatio-temporal data can be significantly enhanced, which in turn improves the performance of STFMs. Furthermore, due to the unique characteristics of spatio-temporal data such as spatial and temporal dependencies, which differ fundamentally from other data types (\emph{e.g.}, language data), embedding techniques play a critical role in aligning data with STFMs. These techniques effectively bridge the gap between raw spatio-temporal data and model input representations.

The second step of STFM pipelines is to \textbf{construct models} based on diverse data. To mitigate the confusion brought by the coarse-grained data type or methodology taxonomies, as shown in the middle of Fig.~\ref{pipeline_taxonomy}, we present a data property taxonomy on STFMs. At the top of our taxonomy, STFMs are classified into two main categories: \emph{primitive} and \emph{transferred} models. This classification is based on whether the models are trained directly on primitive spatio-temporal data or transferred from models that are pre-trained on other data, such as text-based language models or image-based vision models. Furthermore, we divide not only primitive models into temporal, spatial, and spatio-temporal classes according to clear data dependencies, but also transferred models into vision, language, and multi-modal classes based on the available data modalities. The data property taxonomy provides efficient and effective model design and selection because deep learning methods used for the same category under our taxonomy are based on the same data sources, dependencies, or modalities and can be extended to other data types.

The third stage of STFM pipelines delves into \textbf{training objectives} for primitive models and \textbf{adaptation techniques} for transferred models, as shown in the top of Fig.~\ref{pipeline_taxonomy}. Through an in-depth analysis of these methodologies, we highlight their respective advantages and challenges across different data types, tasks, or application scenarios. In the final stage of the pipeline, we examine the \textbf{current applications} of STFMs, showcasing their broad real-world impact across domains such as energy, finance, weather, healthcare, transportation, and public services, as illustrated in Fig.~\ref{param}.

By presenting clear and step-by-step pipelines, our survey not only organizes and clarifies the core components of STFMs but also highlights the deeper connections between them, facilitating the rapid and effective deployment of these models. Moreover, as shown in Table~\ref{intro:comp}, existing surveys often overlook key aspects such as essential spatio-temporal data types, training objectives, and adaptation techniques, resulting in an incomplete understanding of STFMs. Our survey addresses this gap by covering the most comprehensive range of these elements, providing a more holistic perspective on available data sources and their corresponding model training and adaptation methods. Finally, we discuss the current challenges facing STFMs and explore future opportunities for advancing the field.

The main contributions of this survey are summarized as follows.
\begin{itemize}
    \item Comprehensive and up-to-date review. Our paper presents the most extensive and current survey of spatio-temporal foundation models (STFMs), covering a broad spectrum of data types, models, training objectives, and adaptation techniques.
    \item Novel property taxonomy. We propose an innovative data property taxonomy that categorizes STFMs from coarse to fine levels based on data sources and their dependencies, enabling more efficient and effective model design and selection.
    \item Novel pipeline-oriented survey. To the best of our knowledge, this is the first survey to examine STFMs from a pipeline perspective, offering the research community a systematic understanding of how these models are developed and why they achieve superior performance.
    \item Identification of future research opportunities. We outline key challenges in applying foundation models to spatio-temporal tasks, with the goal of guiding future research and inspiring the development of more advanced STFMs.
\end{itemize}

The structure of our survey is as follows. We begin by reviewing the crucial data harmonization process in Section~\ref{sec:data}. Following this, Section~\ref{sec:native} focuses on the model design and training objectives of primitive foundation models. Next, in Section~\ref{sec:transfer}, we delve into transferred foundation models, exploring model selection and transfer adaptation. Section~\ref{sec:app} introduces a broad range of applications where spatio-temporal foundation models have demonstrated significant impact. Beyond the pipeline aspects, Section~\ref{sec:oppo} identifies emerging opportunities and open research challenges in the field. Finally, Section~\ref{sec:conclu} provides a conclusion of the key components of our survey.

\section{Data Harmonization}\label{sec:data}
As shown in the bottom of Fig.~\ref{pipeline_taxonomy}, data harmonization is the first step of pipelines to align raw spatio-temporal data with deep learning methods, which have three important components. The left element is preprocessing, which demonstrates data standardization (\emph{e.g.}, noise filtering) and feature extraction (\emph{e.g.}, temporal pattern decoupling) processes. The middle element is embedding, which transforms preprocessed data into low-dimensional numerical representations, including tokenization techniques to break data into smaller units, spatial embeddings to capture spatial relationships, temporal embeddings to encode time-related features, and frequency embeddings to analyze frequency-based characteristics. Moreover, incorporating additional information through retrieval, exogenous data, and multi-modal inputs can further improve the performance of STFMs. The pipelines from preprocessing to embedding generation, enhanced by side information, to create rich and structured representations of data for STFMs.
\subsection{Trajectory Data}
\subsubsection{Data Description}
Spatio-temporal trajectory data, such as vehicle and pedestrian trajectories, refer to sequences of movement locations recorded with specific timestamps by edge devices like mobile phones, which can be formulated as $\mathcal{X}=\{(l_t,\tau_t)|t=1,2,...,N\}\in{\mathbb{R}^{N\times 3}}$, where $l_t=(lng_t,lat_t)$ denotes the location of the $t^{th}$ point, $\tau_t$ represents the recorded time stamp of the $t^{th}$ point, and $N$ denotes the length of the input trajectory~\cite{deng2024learning}.
\subsubsection{Data Preprocessing}
To train a foundation model with high-quality spatio-temporal trajectory data, many data preprocessing methods have been proposed such as trajectory compression, noise filtering, and map-matching. Specifically, UniTraj~\cite{zhu2024unitraj} filters out meaningless short trajectories and unrealistic fast trajectories, and PTR~\cite{wei2024ptr} inserts placeholder tokens into low-sampling trajectories to transform them into a uniform sampling interval. Besides, map-matching techniques~\cite{yang2018fast} are commonly used in many methods~\cite{jiang2023self,wei2024ptr,zhu2024unitraj,liu2024lighttr} to align recorded spatio-temporal trajectories with realistic road network paths. This is necessary since edge devices used for recording trajectories suffer from errors by building obstructions, poor phone service, and receiver noise~\cite{hofmann2007gnss}.
\subsubsection{Data Embedding}
To incorporate spatio-temporal patterns of individual trajectories into foundation models, various spatio-temporal embeddings are used in current methods. KGTS~\cite{chen2024kgts} uses a knowledge graph to bring local spatial information into spatial embeddings, and MMTEC~\cite{lin2023pre} provides an index-fetching technique to add spatial semantic information of road segments into spatial embeddings. RETE~\cite{chen2023teri} leverages learnable Fourier embedding to capture periodicity in distance and time intervals, and UniTraj~\cite{zhu2024unitraj} uses rotary position embedding~\cite{su2024roformer} to maintain spatio-temporal relative positional information of trajectories for capturing patterns. Moreover, PTR~\cite{wei2024ptr} and START~\cite{jiang2023self} introduce time-of-day and day-of-week temporal embeddings to assist foundation models in learning personalized behavior patterns at specific times. With the assistance of powerful spatio-temporal patterns and travel semantics, trajectory foundation models can be universally adapted to diverse downstream tasks.
\subsubsection{Side Data}
To further enhance the understanding of travel semantics in trajectories of foundation models, external text descriptions of locations and POI categories are used as additional information~\cite{wei2024ptr,lin2024ptrajm}. Moreover, geographic features such as road network data can constrain trajectories to valid routes~\cite{fu2020trembr,zhu2024controltraj} and time-related features such as day-of-week can influence commuting patterns in human mobility trajectories~\cite{wei2024ptr}.

\subsection{Event Data}
\subsubsection{Data Description}
Spatio-temporal event data, such as political events and epidemic outbreaks, refer to a series of quintuples $(s, r, o, l, t)$, where $s$ and $o$ are entities, $r$ is a binary relation between $s$ and $o$, $l$, and $t$ indicate the location and time when event $(s, r, o)$ occurs, respectively. Notably, although location information is often ignored by existing event-related models, the quintuple formulation provides a more general representation of an event. Moreover, $t$ is generally a discrete representation of time, such as day, week, and month, which is based on the practical requirements of an application.

\subsubsection{Data Preprocessing}
Spatio-temporal event data are often represented by graph structures to better model the underlying dependencies among entities and relations. For example, ONSEP~\cite{yu2024onsep} uses a temporal knowledge graph (TKG) to structure the time-sequenced events as a series of multi-relational directed graphs denoted as $TKG_t=\{\mathcal{G}_1, \mathcal{G}_2, \cdots, \mathcal{G}_t\}$, where each $\mathcal{G}_i=(\mathcal{V}, \mathcal{R}, \mathcal{E}_i)$ consists of the events in time $i$. Here, $\mathcal{V}$ and $\mathcal{R}$ denote the sets of entities and relation types, respectively, and $\mathcal{E}_i$ contains a set of event quintuples at time stamp $t$. Similarly, Xia et al.~\cite{xia2024chain} explore and incorporate high-order historical dependencies in TKGs into large language models to improve event inference accuracy. Different from these studies that transform the event forecasting problem into a knowledge graph completion task, Deng et al.~\cite{deng2024advances} focus on political event forecasting and instead formulate it as a classification problem. Specifically, they divide features into two categories, \emph{i.e.}, static $\mathcal{S}_l$ and dynamic $\mathcal{X}_{t-k+1:t, l}$, for each location $l$, and aim to learn a classifier $f(\mathcal{S}_l, \mathcal{X}_{t-k+1:t, l})\rightarrow \mathcal{Y}_{t^*,l}$ that maps the input to a binary event vector $\mathcal{Y}_{t^*,l} \in \{0,1\}^{M}$ at the future time stamp $t^*$ for the target location $l$. Here, $\mathcal{S}_l$ denote a set of static features such as population and political ideology, $\mathcal{X}_{t-k+1:t, l}$ are the collection of dynamic features for location $l$ before time stamp $t$ within a historical window, and $M$ indicates the number of possible event types that can occur concurrently, where the dynamic features can include TKGs or other data formats. 

\subsubsection{Data Embedding}
In order to precisely forecast the occurrence of an event, various sources of information, including text content and causal relationships among events, need to be taken into account. For example, DynamicGCN~\cite{deng2019learning} constructs dynamic graphs based on document-based point-wise mutual information, where each node represents a word extracted from an article. These nodes are then encoded with word embedding vectors pre-trained on the Wikipedia database~\cite{mikolov2013distributed}. Recently, numerous methods have incorporated large language models (LLMs) for event forecasting~\cite{ye2024mirai,yu2024onsep,shi2024language,song2024latent,xia2024chain,yuan2024back}. These methods typically encode historical events by prompt engineering techniques and use LLM-derived text embeddings for reasoning and prediction. For example, Yuan et al.~\cite{yuan2024back} utilize an explainable TKG reasoning model, \emph{e.g.}, TLogic~\cite{liu2022tlogic}, to generate reasoning path embeddings between entities $s$ and $o$, and then generate the prediction and explanation by polished or revised prompts. 

\subsubsection{Side Data}
Generally, events can be concisely represented as quintuples $(s, r, o, l, t)$, denoting the entity, relation, location, and time. Such a representation has been commonly employed in existing graph-based event forecasting studies~\cite{deng2022robust,chen2021civil}, as it effectively captures the structural relationships among events. Nevertheless, such representations often lack the rich contextual information needed for accurate forecasting. To address this limitation, researchers incorporate side information, such as retrieved news articles~\cite{yu2024onsep} and generated content~\cite{yuan2024back}, which provide additional context relevant to specific query events. For instance, Yuan et al.~\cite{yuan2024back} devise a prompt-based mechanism to exploit the generative potential of ChatGPT, thereby generating more diverse and coherent context documents in response to a given query event. This enriched context helps improve both the precision and robustness of event forecasting models. 

\subsection{Spatio-Temporal Grid Data}
\subsubsection{Data Description}
Spatio-temporal grid data, such as traffic flow, crime, and weather data, consist of sequential observations over regularly partitioned grids that span cities or larger geographical areas. These data sequences are typically collected from either automated sources (\emph{e.g.}, satellites, sensors) or human-reported statistics. Formally, this type of data can be formulated as $\mathcal{X}\in\mathbb{R}^{L\times H\times W\times C}$, where $x_{t}^{i,j}\in\mathbb{R}^{C}$ indicates the collected features of the grid cell in the $i^{th}$ row and $j^{th}$ column at time stamp $t$. Next, $L$, $H$, $W$, and $C$ denote the length of the sequence, the number of rows in the grid data, the number of columns in the grid data, and the number of features collected per grid cell, respectively.
\subsubsection{Data Preprocessing}
To address the issue of varying data scales in training a foundation model, normalization methods, such as Z-score~\cite{nguyen2023climax} and Min-Max~\cite{yuan2024unist} normalization, are commonly applied to transform features into a fixed range. To reduce computation needs in training foundation models with high spatial resolutions, a patching technique is used to merge some adjacent grids into a patch~\cite{bi2023accurate}. In detail, the original spatial resolution $H\times W$ will be patched into a small feature map with the resolution $\bar{H}\times \bar{W}$, where $\bar{H}=\frac{H}{h}, \bar{W}=\frac{W}{w}$, and $h\times w$ is the size of each patch. Moreover, 3D patching is further adopted on spatio-temporal grid data to merge grids from spatio-temporal dimensions simultaneously~\cite{yuan2024unist}, \emph{i.e.}, the shape $L\times H\times W$ is patched into $\bar{L}\times \bar{H}\times \bar{W}$, where $\bar{L}=\frac{L}{l}$ and the patch size is $l\times h\times w$. Besides, to extract seasonal and trend information, methods like Fast Fourier transform (FFT) and daily flashback techniques are often used, capturing essential temporal patterns that enhance model understanding~\cite{zheng2023diffuflow,yuan2024urbandit}.
\subsubsection{Data Embedding}
Projection layers are typically used to transform numerical spatio-temporal grid data into high-dimensional representations. To preserve the order and proximity of spatio-temporal grid data within Transformers, diverse positional embedding techniques are designed, including temporal embeddings (\emph{e.g.}, learnable embeddings~\cite{nguyen2023climax} or sinusoidal encodings~\cite{yuan2024unist}) and spatial embeddings (\emph{e.g.}, earth-specific positional biases~\cite{bi2023accurate} and ground sample distance embeddings~\cite{reed2023scale}). Moreover, task-specific embeddings are applied to the variability in feature types and dimensions across different datasets and tasks~\cite{zhao2024weathergfm}, ensuring flexibility and alignment with different tasks.
\subsubsection{Side Data}
Spatio-temporal grid tasks incorporate various external features, including temporal features like holiday, time of day, and day of week; environmental features like temperature and wind speed (beyond the primary input); and spatial features like POI category distributions and structural attributes of road networks. These external features are either directly integrated into primitive foundation models or converted into textual descriptions for transferred foundation models~\cite{zheng2023diffuflow,li2024urbangpt}.

\subsection{Video Data}
\subsubsection{Data Description}
Spatio-temporal video data is composed of a series of consecutive images. Each image is formed by regularly spaced pixels with RGB colors, which can be formulated as $\mathcal{X}\in\mathbb{R}^{L\times H\times W\times 3}$, where $x_{t}^{i,j}\in\mathbb{R}^{3}$ indicates the RGB values of the pixel in the $i^{th}$ row and $j^{th}$ column of the $t^{th}$ frame. Next, $L$, $H$, and $W$ denote the number of frames in the video and the spatial resolution (height and width) of each frame, respectively.
\subsubsection{Data Preprocessing}
Similar to spatio-temporal grid data, video data often require patching when processed by models without the pre-trained vision foundation models~\cite{chen2023videollm}, where the patched shape is $\bar{L}\times \bar{H}\times \bar{W}$ with the patch size $l\times h\times w$. Notably, video data can be directly fed into pre-trained vision foundation models, \emph{e.g.}, ViT~\cite{dosovitskiy2020image}, without any preprocessing operations~\cite{li2023videochat,song2024moviechat}.
\subsubsection{Data Embedding}
For foundation models that do not use pre-trained vision backbones, pre-trained tokenization models, such as VQ-GAN and VQ-VAE, should be utilized on the video data to transform raw pixels into discrete embedded tokens~\cite{chen2024seine,gu2024seer}. Alternatively, pre-trained image models, such as CLIP ViT-L/14~\cite{radford2021learning}, are leveraged to process each video frame independently instead of using pre-trained video models to embed pixels into visual features with temporal semantics~\cite{yang2022zero,pan2023retrieving}. To obtain LLM-compatible embeddings for video understanding tasks, models like pre-trained Q-former from BLIP~\cite{li2023blip} should be applied on each embedded frame followed by pre-trained image models~\cite{li2023videochat,song2024moviechat}. Moreover, temporal positional embeddings can be added to frame-level representations after embedding the video data since temporal dependencies are often ignored in pre-trained tokenization, vision, and multi-modal models~\cite{zhang2023video}.
\subsubsection{Side Data}
Video-text pairs from text corpora are essential external factors for multi-modal video tasks, such as video question answering and video understanding~\cite{yang2022zero,pan2023retrieving}. For single-modality-based prediction tasks, captions of each frame should be automatically derived through pre-trained vision-language models~\cite{yang2024generalized}. Moreover, the audio modality and future action conditions are often considered as auxiliary information to support video-based tasks~\cite{zhang2023video,yang2024generalized}.

\subsection{Spatio-Temporal Graph Data}
\subsubsection{Data Description}
Spatio-temporal graph data, such as traffic speed, electroencephalography (EEG), and electricity data, refer to sequential observations recorded by irregularly distributed sensors in a spatial modeled as graphs. This type of data can be defined as $\mathcal{X}\in\mathbb{R}^{L\times N\times C}$, where $x_t^n\in\mathbb{R}^{C}$ denotes the feature vector recorded by sensor $n$ at time stamp $t$. Next, $L$, $N$, and $C$ denote the length of the sequence, the number of observations, and the number of features for each observation. The key difference between spatio-temporal grid data and graph data lies in how spatial correlations are defined. In graph data, spatial correlations are manually defined by various spatial metrics rather than relying on explicitly adjacent grid cells, as is the case with grid data.
\subsubsection{Data Preprocessing}
For learning intricate temporal patterns of spatio-temporal graph data, time series decoupling techniques such as wavelet decomposition and seasonal-trend decomposition procedure based on loess are applied on spatio-temporal graph data to isolate different temporal patterns and avoid mutual influence~\cite{zhang2024score,yangfrequency,zhang2024brant}. Moreover, long spatio-temporal graph sequences are typically selected as input for most STFMs~\cite{shao2022pre,li2023generative,gao2024spatial}. The input sequence with length $L$ is segmented into $\bar{L}$ non-overlapping patches to decrease the complexity of the foundation models, because the information density of original spatio-temporal graph sequences is lower than texts, where the temporally patched sequences can be mathematically formulated as $\bar{L}=\frac{L}{l}$, and $l$ denotes the capacity of each patch. Moreover, to efficiently model the spatial dimension of graph data, an irregular spatial patching technique has been proposed~\cite{fang2024efficient}. Specifically, indices of $N$ observations are reordered so that spatially adjacent observations are arranged consecutively, allowing $N$ to be divided into $\bar{N}$ non-overlapping patches. This segmentation decreases the complexity of the spatial dimension, where the spatially patched sequences can be mathematically formulated as $\bar{N}=\frac{N}{n}$, where $n$ denotes the capacity of each patch.
\subsubsection{Data Embedding}
To further enable the foundation models to grasp structural and periodic characteristics of spatio-temporal graph data, various temporal and spatial embeddings have been proposed and integrated as part of the model inputs. For example, the learnable point embeddings~\cite{shao2022pre,li2023generative}, fixed point embeddings (\emph{e.g.}, day-of-week embeddings~\cite{li2024flashst,li2024opencity,gao2024spatial}), and power spectral density-based frequency embeddings~\cite{yuan2024ppi,zhang2024brant} have been successively proposed as temporal embeddings, serving as one of the model's input components. In terms of spatial embeddings, sinusoidal embeddings~\cite{gao2024spatial}, distance embeddings~\cite{wu2024g2ptl}, and eigenvectors-based graph spectral embeddings~\cite{li2024opencity} are utilized to inject essential spatial structure information into foundation models. By incorporating these spatial, temporal, frequency, and spectral embeddings, spatio-temporal graph foundation models gain a deeper understanding of the structural and periodic nature of spatio-temporal graph data, significantly enhancing their generalization capabilities across diverse tasks and domains.
\subsubsection{Side Data}
Similar to spatio-temporal grid tasks, spatio-temporal graph tasks also incorporate temporal features like holiday, time of day, and day of week and spatial factors like POI category distributions and structural attributes of road networks. Moreover, as weather is correlated with wind power generation, temperature and humidity are often used as exogenous variables in spatio-temporal graph-based power foundation models~\cite{tu2024powerpm}.

\begin{table*}[ht]
    \aboverulesep=0ex
    \belowrulesep=0ex
    \begin{threeparttable}
    \caption{Summary of used datasets in STFMs. }
    \label{data:datasets}
    \centering
    \begin{tabular}{c|cccccc}
    \toprule
        \multicolumn{1}{c|}{Domains} & \multicolumn{1}{c}{Datasets} & \multicolumn{1}{c}{Spatial Size} & \multicolumn{1}{c}{Temporal Range} & \multicolumn{1}{c}{Applications} & \multicolumn{1}{c}{Sampling Interval} & \multicolumn{1}{c}{Features}\\
    \midrule
        \multirow{9}{*}{Trajectory} & Chengdu~\cite{zhu2024unitraj} & 3,032,212 & 10/01/2018-11/30/2018 & Transportation & irregular & Taxi\\
        & Xian~\cite{zhu2024unitraj} & 18,267,443 & 10/01/2018-11/30/2018 & Transportation & irregular & Taxi\\
        & Porto~\cite{zhu2024unitraj} & 12,852,750 & 07/01/2013-11/01/2013 & Transportation & irregular & Taxi\\
        & LaDe~\cite{wu2023lade} & 10,677,000 & 6 months & Transportation & irregular & Delivery\\
        & TKY~\cite{yang2014modeling} & 537,703 & 04/04/2012-02/16/2013 & Transportation & irregular & Check-in\\
        & NYC~\cite{yang2014modeling} & 227,428 & 04/03/2012-02/15/2013 & Transportation & irregular & Check-in\\
        & Gowalla~\cite{cho2011friendship} & 6,442,892 & 11/2010-12/2011 & Transportation & irregular & Check-in\\
        & GeoLife~\cite{zheng2010geolife} & 23,667,828 & 04/2007-10/2011 & Transportation & irregular & Mobile\\
        & WorldTrace~\cite{zhu2024unitraj} & 8.8 Billion & 08/2021-12/2023 & Transportation & 1 second & OSM\tnote{1}\\
    \midrule
        \multirow{5}{*}{Event} & OpenForecast~\cite{wang2025openforecast} & 516,572 & 1950-2024 & Public Services & irregular & News\\
        & GDELT~\cite{leetaru2013gdelt} & 200 Million & 1979-2012 & Public Services & irregular & News\\
        & ICEWS14~\cite{zhang2022along} & 7,128 & 2014 & Public Services & 1 day & Political\\
        & Complaint~\cite{zhang2023promptst} & 2.27 Million & 2013-2014 & Public Services & irregular & Complaint\\
        & MIMIC~\cite{johnson2016mimic} & 53,423 & 2001-2012 & Healthcare & irregular & ICU\\
    \midrule
        \multirow{13}{*}{Spatio-Temporal Grid} & NYC-TAXI~\cite{li2024opencity} & 263 & 01/01/2016-12/31/2021 & Transportation & 30 minutes & Demand\\
        & CHI-TAXI~\cite{li2024opencity} & 77 & 01/01/2021-12/31/2021 & Transportation & 30 minutes & Demand\\
        & NYC-BIKE~\cite{li2024opencity} & 2162 & 01/01/2016-12/31/2021 & Transportation & 30 minutes & Flow\\
        & TrafficSH~\cite{li2024opencity} & 28 $\times$ 32 & 03/05/2022-04/05/2022 & Transportation & 30 minutes & Speed\\
        & TrafficCD~\cite{li2024opencity} & 26 $\times$ 28 & 03/05/2022-04/05/2022 & Transportation & 30 minutes & Speed\\
        & WeatherBench~\cite{rasp2020weatherbench} & 32 $\times$ 64 & 1979-2018 & Weather & 6 hours & ERA5\tnote{2}\\
        &  SEVIR~\cite{veillette2020sevir} & 384 $\times$ 384 & 2017-2019 & Weather & 5 minutes & Storm\\
        & CMIP6-ClimaX~\cite{nguyen2023climax} & - & 1850-2015 & Weather & 6 hours & CMIP6\tnote{3}\\
        & SST~\cite{jamali2022satellite} & 60 $\times$ 60 & 1982-2021 & Weather & 1 day & Temperature\\
        & NYC-Crime~\cite{li2022spatial} & 2162 & 01/01/2016-12/31/2021 & Public Services & 1 day & Crime\\
        & LA-Crime~\cite{wang2024diffcrime} & 16 $\times$ 16 & 2017-2022 & Public Services & 1 day & Crime\\
        & CHI-Crime~\cite{li2022spatial} & 168 & 01/2016-12/2017 & Public Services & 1 day & Crime\\
        & PopSH~\cite{yuan2024urbandit} & 32 $\times$ 28 & 08/01/2014-08/28/2014 & Public Services & 1 hour & Population\\
    \midrule
        \multirow{8}{*}{Video} & KITTI~\cite{geiger2012we} & $128\times 160$ & 5 days & Transportation & 100 millisecond & Driving\\
        & KTH~\cite{schuldt2004recognizing} & $120\times 160$ & 4 seconds & Public Services & 40 millisecond & Action\\
        & UCF101~\cite{soomro2012ucf101} & $320\times 240$ & 7 seconds & Public Services & 40 millisecond & Action\\
        & Avenue~\cite{lu2013abnormal} & - & 2 minutes & Transportation & 7 millisecond & Campu\\
        & SSV2~\cite{goyal2017something} & $84\times 84$ & 4 seconds & Public Services & 42 millisecond & Interaction\\
        & WebVid~\cite{bain2021frozen} & $224\times 224$ & 18 seconds & Public Services & - & MultiModal\\
    \midrule
        \multirow{21}{*}{Spatio-Temporal Graph} & METR-LA~\cite{li2017diffusion} & 207 & 03/01/2012-06/30/2012 & Transportation & 5 minutes & Speed\\
        & PEMS-BAY~\cite{li2017diffusion} & 325 & 01/01/2017-05/31/2017 & Transportation & 5 minutes & Speed\\
        & PEMS03~\cite{song2020spatial} & 358 & 09/01/2018-11/30/2018 & Transportation & 5 minutes & Flow\\
        & PEMS04~\cite{song2020spatial} & 170 & 01/01/2018-02/28/2018 & Transportation & 5 minutes & Flow\\
        & PEMS07~\cite{song2020spatial} & 307 & 05/01/2017-08/31/2017 & Transportation & 5 minutes & Flow\\
        & PEMS08~\cite{song2020spatial} & 325 & 07/01/2016-08/31/2016 & Transportation & 5 minutes & Flow\\
        & LargeST~\cite{liu2023largest} & 8,600 & 01/01/2017-12/31/2021 & Transportation & 15 minutes & Flow\\
        & OSAP~\cite{fang2024spatio} & 1015 & 1970-2019 & Financial & 1 day & Stock\\
        & Exchange~\cite{wu2020connecting} & 8 & 1990-2016 & Financial & 1 day & Exchange\\
        & IGM~\cite{wimmer2023leveraging} & 1 & 04/21/2020-04/20/2021 & Financial & 1 day & Market\\
        & AIR-BJ~\cite{yi2018deep} & 34 & 05/2014-04/2017 & Weather & 1 hour & Air\\
        & AIR-GZ~\cite{yi2018deep} & 41 & 05/2014-04/2017 & Weather & 1 hour & Air\\
        & AIR-China~\cite{liang2023airformer} & 1085 & 01/01/2015-12/31/2018 & Weather & 3 hours & Air\\
        & Brant~\cite{zhang2024brant} & 134 & 700 hours & Healthcare & - & EEG\\
        & SEED~\cite{zheng2015investigating} & 62 & - & Healthcare & 1 millisecond & EEG\\
        & BCI~\cite{blankertz2007non} & 59 & - & Healthcare & 1 millisecond & EEG\\
        & PhysioNet~\cite{goldberger2000physiobank} & 11,988 & 48 hours & Healthcare & - & ICU\\
        & Solar-Energy~\cite{wu2020connecting} & 137 & 2007 & Energy & 10 minutes & Solar\\
        & Electricity~\cite{wu2020connecting} & 321 & 2012-2014 & Energy & 1 hour & Electric\\
        & CAISO~\cite{tu2024powerpm} & 34 & 04/25/2023-04/23/2024 & Energy & 1 hour & Electric\\
        & PJM~\cite{tu2024powerpm} & 22 & 03/28/2024-04/26/2024 & Energy & 5 minutes & Electric\\
    \bottomrule
    \end{tabular}
    \begin{tablenotes}
        \item[1] https://wiki.openstreetmap.org/
        \item[2] https://cds.climate.copernicus.eu/datasets/reanalysis-era5-single-levels
    \end{tablenotes}
    \end{threeparttable}
\end{table*}
\subsection{Datasets}
The emergent abilities of general-purpose models are heavily influenced by the size and quality of the training data. This suggests that as the model encounters more diverse and high-quality data, it becomes more capable of generalizing across a wide range of tasks. For spatio-temporal models, which capture patterns over both space and time, the diversity and richness of the data are crucial for effectively identifying complex spatio-temporal relationships. As illustrated in Table~\ref{data:datasets}, we summarize the trajectory, event, spatio-temporal grid, video, and spatio-temporal graph datasets adopted across different spatio-temporal foundation models, including the spatial size, temporal range, sampling interval of the temporal dimension, and the recorded features of data. These datasets are widely used in various real-world applications, \emph{e.g.}, transportation, weather, healthcare, energy, financial, and public services. As shown in Table~\ref{data:datasets}, one of the significant challenges highlighted is the variability across datasets, even within the same application domain (\emph{e.g.}, transportation or healthcare). For example, spatial size, temporal range, and sampling intervals can differ widely, making it difficult for a single foundation model to generalize across these differences. This could lead to issues where a model trained on one dataset performs poorly on another due to the mismatched characteristics of the data. The heterogeneity of the datasets adds complexity to model training and deployment, as models may struggle to adapt to different scales and granularities of data. Another challenge is the comparison between the scale of spatio-temporal datasets and that of LLMs. Spatio-temporal datasets tend to be much smaller in scale, often in the million-item range, compared to LLMs, which can be trained on datasets with billions of items. This suggests that while LLMs benefit from massive amounts of data, spatio-temporal foundation models may have limited scalability due to the smaller size and specialized nature of the datasets available. The fact that many smaller datasets are used to train these models indicates that scaling them up to the same size as LLMs may not be feasible, or at least may not yield the same performance improvements. On the other hand, the statement about scaling laws also hints at a potential trade-off between data quality and quantity for spatio-temporal foundation models. While increasing the quantity of data might improve a model's performance, the effectiveness of spatio-temporal models might be more dependent on the quality and relevance of the data rather than sheer size. This is particularly important in fields like healthcare, weather, or transportation, where specific domain knowledge and high-quality data may matter more than the raw scale. In summary, the heterogeneity of spatio-temporal data poses challenges, and while scaling these models is important, it's likely that other factors such as data quality and domain relevance will play a more significant role in improving performance than only increasing dataset size.

\section{Primitive Spatio-Temporal Foundation Models}\label{sec:native}
As an essential branch of spatio-temporal foundation models, \emph{primitive} models refer to foundation models pre-trained with primitive spatio-temporal data, which can be generalized across different downstream tasks. To derive primitive spatio-temporal foundation models, as illustrated in Fig.~\ref{pipeline_taxonomy}, model architectures need to be carefully designed based on spatio-temporal properties of the embedded data. Then the presented primitive STFM is trained with specific self-supervised learning objectives to achieve general capabilities according to the input data type and the applied downstream task. We present the detailed model design and objectives below.

\subsection{Model Design}
In this study, we present the model design of STFMs according to three categories: temporal models, spatial models, and spatio-temporal models. For temporal models, spatial dependencies are not taken into account due to insignificant spatial distinguishability of spatio-temporal data~\cite{shao2024exploring}. For spatial models, temporal dependencies are not taken into account because of unclear and unstable temporal patterns of spatio-temporal data~\cite{shao2024exploring}. For spatio-temporal models, spatio-temporal dependencies are simultaneously needed because of the significant spatial distinguishability and temporal patterns of the input data.
\subsubsection{Temporal Models}
A key aspect of primitive spatio-temporal foundation models is to capture the complex temporal dependencies, such as trend and seasonality~\cite{fang2023spatio}. Specifically, to model short-term temporal dependencies, recurrent neural networks, such as GRU~\cite{shu2021short} and LSTM~\cite{bogaerts2020graph}, are often adopted, especially for spatio-temporal trajectory and graph data~\cite{deng2022efficient,ma2024more,yuan2024ppi}. However, recurrent neural networks fall short in capturing long-term temporal dependencies~\cite{zhou2022fedformer}. To address this problem, Vanilla Transformer is utilized in temporal-only foundation models, \emph{e.g.}, TrajFM~\cite{lin2024trajfm}, UniTraj~\cite{zhu2024unitraj}, EEGPT~\cite{yue2024eegpt}, and STD-MAE~\cite{gao2024spatial}, to model long-term temporal dependencies. In particular, patch embedding layers are often used before Transformers for long-term spatio-temporal graph data~\cite{shao2022pre,jiang2024large}, with the aim of reducing computation costs. Further, the pure MLP architecture is adopted in TTM~\cite{ekambaram2024tiny} to substitute Transformers for efficient long-term time series modeling. In terms of efficient trajectory modeling, the neural controlled differential equation is proposed in MMTEC~\cite{lin2023pre}. In addition to the above-mentioned discriminative foundation models, diffusion model based methods~\cite{ho2020denoising} have also been proposed for generative tasks, \emph{e.g.}, ControlTraj~\cite{zhu2024controltraj} and Score-CDM~\cite{zhang2024score}. For conditional temporal-only diffusion models, the attention mechanism is widely used in the denoising networks to learn temporal patterns from conditional input. For example, ControlTraj~\cite{zhu2024controltraj} and FGTI~\cite{yangfrequency} adopt geo-attention and time-frenquency attention to learn road constraints in trajectory generation and frequency knowledge in spatio-temporal graph imputation, respectively. Besides, frequency kernel supported global convolution is proposed as the denoising network in Score-CDM~\cite{zhang2024score} to replace attention for spatio-temporal graph imputation.

\subsubsection{Spatial Models}
Topology-constrained spatio-temporal data often exhibit complex spatial dependencies~\cite{wu2019graph}, including local and global spatial correlations~\cite{fang2023spatio}. Local spatial correlations are always reflected by the first law of geography, \emph{i.e.}, things that are near are more related than things that are far away. Moreover, global spatial correlations also exist in spatio-temporal data due to the regional functionality. Thus, another line of research focuses on capturing spatial correlations by means of foundation models for spatio-temporal data. To model local spatial dependencies, existing studies often use graph neural networks in various tasks~\cite{zhang2023automated,zhang2023spatial} with a pre-defined topology (\emph{e.g.}, road networks and paths)~\cite{chang2023spatial}. However, it is difficult to fully pre-define the complex spatial dependencies by spatial topologies, where global spatial correlations are often ignored~\cite{wu2019graph,wu2020connecting}. To address this problem, Vision Transformers are utilized in foundation models to capture global spatial dependencies of spatio-temporal grid data~\cite{nguyen2023climax,zhao2024weathergfm,chen2023fengwu,reed2023scale,yuan2022contextualized}. Further, UniST~\cite{yuan2024unist} and PanGu-Weather~\cite{bi2022pangu} apply the Vision Transformer on the 3D-patched grid data to capture cross-time spatial dependencies. Moreover, FourCastNet~\cite{pathak2022fourcastnet} leverages the Fourier neural operator to substitute the attention mechanism in Transformers for efficient grid data-based global spatial correlations modeling. To combine the advantages of graph neural networks and Transformers, G2PTL~\cite{wu2024g2ptl} integrates the graph Transformer into the geography foundation model, where the spatial correlation matrix calculated by attention is refined by a pre-defined spatial topology and thus a more realistic matrix is derived. Moreover, generative foundation models with only spatial diffusion are also explored for crime risk inference~\cite{wang2024diffcrime} and trajectory generation~\cite{zhu2023difftraj}. Specifically, the concatenation-based conventional U-Net~\cite{wang2024diffcrime} and the cross-attention enhanced U-Net~\cite{zheng2023diffuflow} are used in the diffusion model to capture conditional spatial knowledge of grid data effectively.

\subsubsection{Spatio-Temporal Models}
Spatio-temporal models aim to learn intricate spatial and temporal correlations simultaneously~\cite{li2024opencity,tu2024powerpm}. Orginally, the trajectory foundation model KGTS~\cite{chen2024kgts} combines recurrent neural networks and graph neural networks to learn temporal and spatial correlations, respectively. Recently, to efficiently capture long-term temporal dependencies, temporal convolutions and Transformers emerge, which are combined with graph neural networks to build foundation models across spatio-temporal data~\cite{ji2023spatio,jiang2023self,li2024opencity,deng2024multi}. On the spatial aspect, the traditional graph neural networks with limited local receptive fields are also upgraded by hypergraph neural networks, hierarchical graph neural networks, and Transformers in spatio-temporal graph and grid tasks~\cite{li2022spatial,tu2024powerpm,zhang2024brant} due to their capabilities to capture global spatial correlations. Moreover, MSTEM~\cite{gu2024mstem} integrates all of the spatio-temporal convolutions and spatio-temporal Transformers into one foundation model to learn long-short spatio-temporal knowledge for event forecasting. For generative conditional diffusion models, spatio-temporal convolutions and Transformers are often chosen as the denoising network for historically-concatenated spatio-temporal graph data~\cite{wen2023diffstg,hu2024towards}. Moreover, in addition to spatio-temporal Transformers, spatio-temporal cross-attention is used on unmasked trajectories in the traffic signal control model DiffLight~\cite{chen2024difflight} to capture spatio-temporal condition information.

\subsection{Training Objectives}
Training objectives are also crucial for primitive STFMs, and aim to guide the direction of model updates for achieving satisfactory results. We present four promising training objectives in STFMs. We begin with \emph{regression modeling}, exploring how it learns universal historical knowledge. Then, we study \emph{masked modeling}, pointing out the importance of learning knowledge from bidirectional spatio-temporal contexts. Next, we focus on \emph{contrastive learning}, emphasizing its role in learning generalizable representations of spatio-temporal data. Finally, we introduce the \emph{diffusion generation}, revealing its potential for generative pre-training.

\begin{figure}[ht]
  \centering
\includegraphics[width=0.8\linewidth]{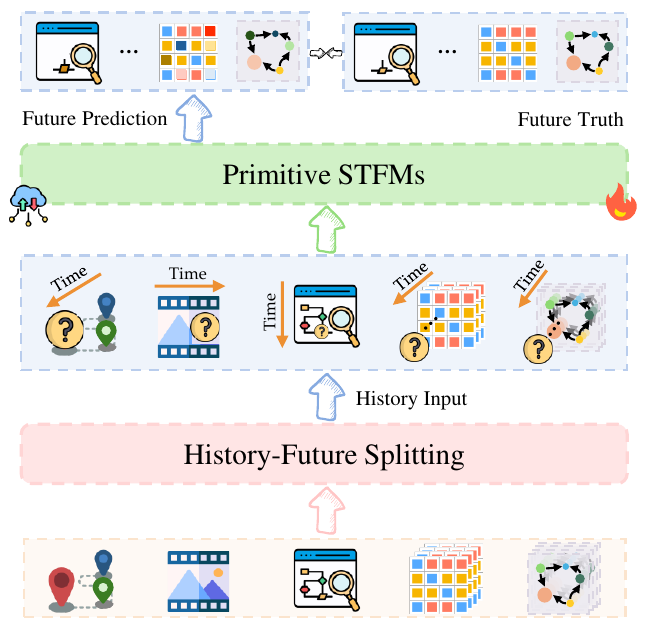}
  \caption{The regression modeling objective, where \includegraphics[scale=0.025]{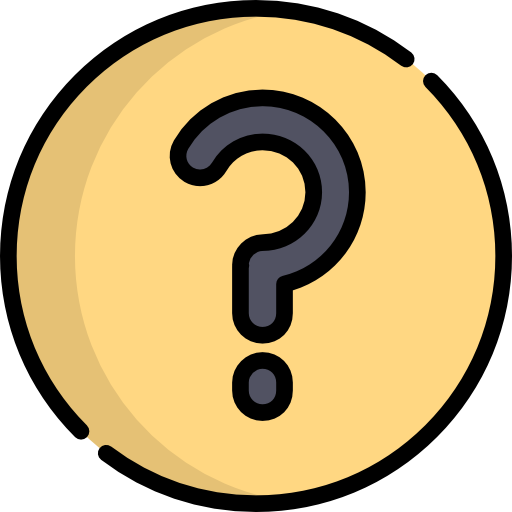} and \includegraphics[scale=0.07]{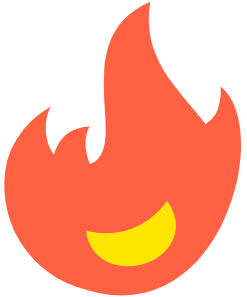} indicate the unknown future observation and active parameters, respectively.}
  \label{ssl:reg}
\end{figure}
\subsubsection{Regression Modeling}
Inspired by the success of autoregressive trained large language and vision models~\cite{du2024survey}, regression has become a powerful self-supervised optimization function during the pre-training stage based on the sequence properties of spatio-temporal data. The regression paradigm is shown in Fig.~\ref{ssl:reg}, where STFMs are trained by the historical ST data and aim to forecast the future.

For example, PointGPT~\cite{chen2024pointgpt} directly forecasts point patches autoregressively during the pre-training phase for point cloud-based tasks. To achieve a universal trajectory model, UVTM~\cite{lin2024gtm} and TrajFM~\cite{lin2024trajfm} support different trajectory-related tasks, which autoregressively generate future features with the mask placeholder. Moreover, FourCastNet~\cite{pathak2022fourcastnet} iteratively utilizes the output of the trained high-resolution weather forecasting model as the input of the next timestamp in the inference stage. Apart from autoregressive training, OpenCity~\cite{li2024opencity} trains a large foundation model for traffic forecasting with the regression function, which aims to perform forecasting only once. In addition, FengWu~\cite{chen2023fengwu} treats the numerical weather prediction as a multi-task regression problem based on the view that the prediction of each variable can be treated as a independent task. ClimaX~\cite{nguyen2023climax} proposes a randomized forecasting objective for foundation model training, where the goal is to predict an arbitrary set of input variables at an arbitrary time into the future. LaBraM~\cite{jianglarge} pre-trains a neural tokenizer model through predicting the Fourier spectrum of the EEG time series. Further, the EEGPT~\cite{yue2024eegpt} and iVideaoGPT~\cite{wu2024ivideogpt} perform next-token prediction on the EEG and video data, which transform the sequential input data into tokens via embedding techniques. Besides, the next-frame and next-scale predictions are also explored in video-related tasks~\cite{hudson2024everything,tian2024visual}. However, most existing regression-based methods rely on the iterative prediction paradigm, which requires significant computational resources and fails to adapt to streaming settings. It is promising to invent new methods to improve efficiency and mitigate time series distribution shifts in the future.

\begin{figure}[ht]
  \centering
\includegraphics[width=0.8\linewidth]{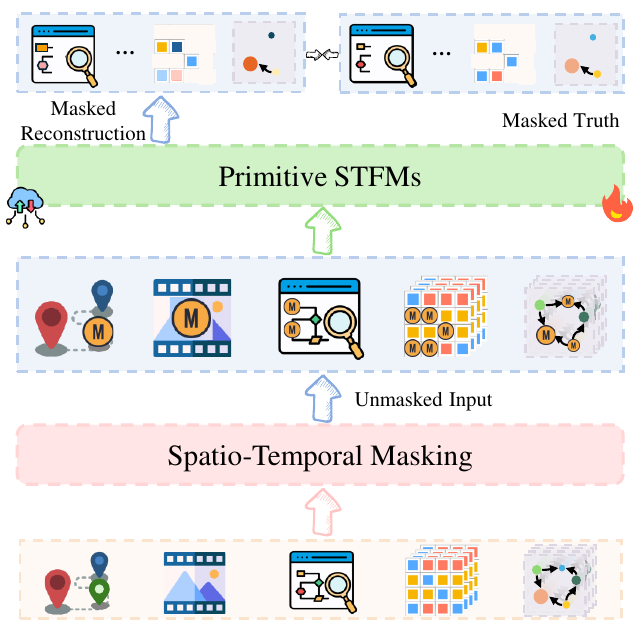}
  \caption{The masked modeling objective, where \includegraphics[scale=0.025]{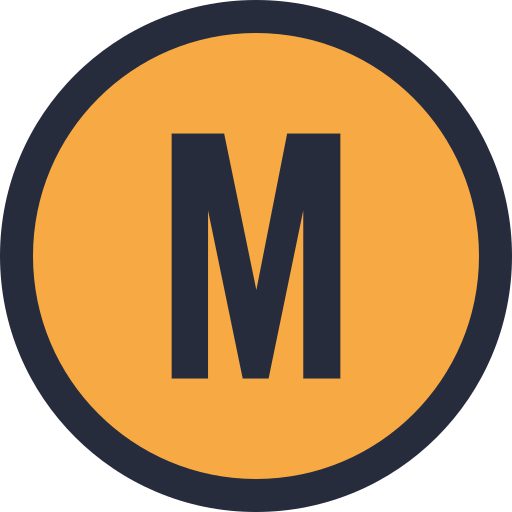} and \includegraphics[scale=0.07]{pics/fire.png} indicate the masked observation and active parameters, respectively.}
  \label{ssl:masked}
\end{figure}
\subsubsection{Masked Modeling}
Benefiting from the success of masked language and image modeling, \emph{e.g.}, BERT~\cite{kenton2019bert} and masked autoencoder (MAE)~\cite{he2022masked}, masked modeling has emerged as a significant pre-training strategy for spatio-temporal foundation models~\cite{feichtenhofer2022masked,sun2023revealing,tong2022videomae,cai2023marlin}. As shown in Fig.~\ref{ssl:masked}, the conventional MAE randomly masks tokens, allowing its encoder to learn from unmasked tokens and the decoder to recover masked ones.

At the beginning, STEP~\cite{shao2022pre} uses the conventional MAE on the temporal dimension of spatio-temporal graph data to pre-train a foundation model for learning the robust representation of long input, which can be used to construct dynamic graphs in traffic forecasting. In addition, GPT-ST~\cite{li2023generative} uses cluster-aware masking to replace the random masking in STEP for cross-cluster knowledge learning. However, both of them only pre-trained an auxiliary tool instead of a model that can be directly applied to downstream tasks. Therefore, Brant~\cite{zhang2024brant} masks the neural signal to train a spatio-temporal encoder-only intracranial neural signal foundation model for analyzing a broad range of tasks. PowerPM~\cite{tu2024powerpm} follows this paradigm to pre-train a power foundation model with a hierarchical spatial encoder. Next, UniTraj~\cite{zhu2024unitraj} proposes block masking and key points masking on the trajectory data to achieve a trajectory foundation model. Different from the above methods that mask tokens on the temporal dimension, Scale-MAE~\cite{reed2023scale}, G2PTL~\cite{wu2024g2ptl} and WeatherGFM~\cite{zhao2024weathergfm} pre-train foundation models by masking the spatial dimension of grid data, which can handle various downstream tasks in terms of geographic and weather. Further, Point-MAE~\cite{pang2022masked} uses the MAE on the patched tokens of 3D point cloud to train a point cloud foundation model, TFMAE~\cite{fang2024temporal} masks time series with small amplitude in the frequency domain, and EMSTGAE~\cite{yu2023ensembled} leverages the graph masking strategy for the spatio-temporal graph data. Consequently, UniST~\cite{yuan2024unist} combines spatial and temporal masking to pre-train a foundation model related to spatio-temporal grid tasks across different applications. Moreover, STD-MAE~\cite{gao2024spatial} reveals that using two separate spatial and temporal MAEs during the pre-train stage can achieve better performance in traffic forecasting. However, most existing masked modeling methods fail to contend with distribution shifts when meeting streaming spatio-temporal data, resulting in significant performance degradation. In addition, current methods are mostly used on high-density information data such as spatio-temporal graph data~\cite{shao2022pre}, and the performance of using the masked technique on spatio-temporal data with low-density information such as event data needs to be further improved in the future.

\begin{figure}[ht]
  \centering
\includegraphics[width=0.8\linewidth]{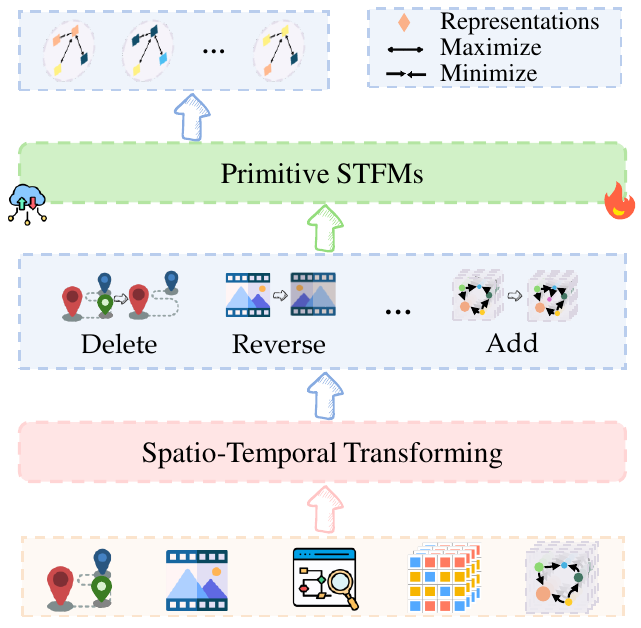}
  \caption{The contrastive training objective, where \includegraphics[scale=0.07]{pics/fire.png} indicates active parameters.}
  \label{ssl:cl}
\end{figure}
\subsubsection{Contrastive Learning}
Contrastive learning~\cite{NEURIPS2020_d89a66c7} exhibits strong capabilities in aligning representations of different views of data to improve performance, which has been verified in vision~\cite{yang2022vision}, language~\cite{diba2021vi2clr}, and graph~\cite{you2020graph} tasks. Contrastive learning is also a mainstream pre-training strategy in spatio-temporal data mining~\cite{ding2022dual,yuan2022contextualized,tang2023spatio,gao2021efficient,fang2023stwave+}, and the process of contrastive learning is shown in Fig.~\ref{ssl:cl}.

STGCL~\cite{liu2022contrastive} first explores the effectiveness of contrastive pre-training in traffic forecasting, and designs four spatio-temporal data augmentation strategies (edge masking and input masking on the spatial domain, and temporal shifting and input smoothing on the temporal domain) to derive different views of data as the positive sample for contrastive learning. STSSL~\cite{ji2023spatio} and SPGCL~\cite{li2022mining} further propose an adaptive spatio-temporal graph augmentation strategy to avoid the hand-craft issue. Using contrastive learning differently on the same representation granularity of the data, CL-TSim~\cite{deng2022efficient} and TrajCL~\cite{chang2023contrastive} incorporate contrastive learning into route representation learning and trajectory similarity learning, respectively, with diverse point- (distorting and masking) and trajectory-level (truncating and simplification) augmentation strategies. To perform spatio-temporal graph contrastive learning on different granularities, ST-HSL~\cite{li2022spatial} considers representations achieved from convolution and hypergraph convolution as different views for effective spatio-temporal modeling. In particular, the convolution and hypergraph convolution in ST-HSL can collaboratively enhance each other to mitigate data sparsity, which is important for various ST tasks, \emph{e.g.,} crime prediction. However, these methods mainly focus on one specific downstream task and are hard to generalize to other domains or tasks. To address this problem, AutoST~\cite{zhang2023automated} proposes an automated spatio-temporal graph contrastive learning paradigm to adaptively derive different views. The automated contrastive paradigm on the heterogeneous graph neural network is implemented by a random walk with learnable Gaussian noise, which alleviates noise and distribution heterogeneity. FlashST~\cite{li2024flashst} uses contrastive learning on the prompt embedding to ensure a consistent distribution across diverse downstream tasks based on an optimized uniform embedding distribution. To learn a universal trajectory representation for downstream tasks, START~\cite{jiang2023self}, JGRM~\cite{ma2024more}, and MMTEC~\cite{lin2023pre} utilize the map-matched road network as a positive view to introduce travel semantics into the foundation model. PTrajM~\cite{lin2024ptrajm} further uses the closest POI to each trajectory point as another positive POI view to enhance travel semantics. To deal with the multi-modal urban spatial image and text description data, UrbanCLIP~\cite{yan2024urbanclip} designs an urban foundation model with language-image contrastive pre-train. Specifically, representations from different modals that have the same semantics are aligned by contrastive learning in UrbanCLIP, and the text-enriched model achieves significant improvements in downstream tasks. Despite the fact that contrastive learning can mitigate distribution shifts according to the consistency of different views of the same input~\cite{fang2024temporal}, it still requires additional computing resources to fine-tune projection heads for adapting specific downstream tasks.

\begin{figure}[ht]
  \centering
\includegraphics[width=0.8\linewidth]{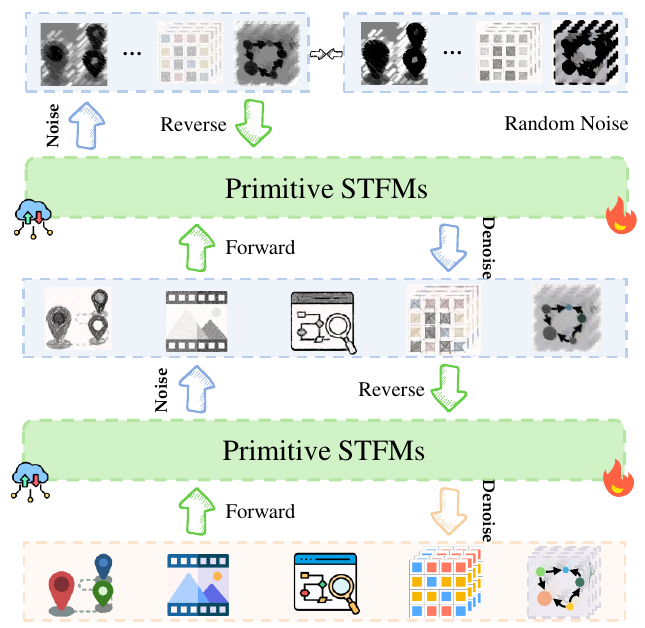}
  \caption{The diffusion generation objective, where \includegraphics[scale=0.07]{pics/fire.png} indicates active parameters.}
  \label{ssl:dd}
\end{figure}
\subsubsection{Diffusion Generation}
Denoising diffusion models have achieved prominence in unsupervised visual generation tasks~\cite{rombach2022high,peebles2023scalable} due to their abilities to learn distributions of intricate unlabeled data, which can adapt to unsupervised spatio-temporal grid and video tasks~\cite{xu2023diffusion,zhang2024score,chen2024difflight,ruhling2024dyffusion}. As shown in Fig.~\ref{ssl:dd}, denoising diffusion models first add Gaussian noise to transform data (forward process) into standard Gaussian and then utilize a network to learn data reconstruction from noise (reverse process).

To adapt diffusion models with spatio-temporal forecasting, DiffSTG~\cite{wen2023diffstg} and MaTCHS~\cite{daiya2024diffstock} treat the history data as the condition in the reverse process and concatenate it with the noised data to generate target future. Subsequently, CaPaint~\cite{duan2024causal} utilizes the discovered causal data as the condition to generate multiple sequences. On the other hand, SPDiff~\cite{chen2024social} and DiffCrime~\cite{wang2024diffcrime} utilize two independent channels in the denoising network to learn representations of noised data and the history condition, which can mitigate the negative influence caused by the discrepancy of the data. Moreover, DiffTraj~\cite{zhu2023difftraj} utilizes a diffusion model for trajectory recovery, and the recorded motion properties serve as conditions. In particular, the forecasting task is considered as a special case of recovery in USTD~\cite{hu2024towards}, \emph{i.e.}, the future to be predicted is a missing value to be recovered. To further extract deterministic conditional knowledge, existing studies in imputation, forecasting, and fine-grained inference like DSTPP~\cite{yuan2023spatio}, PriSTI~\cite{liu2023pristi}, DiffUFlow~\cite{zheng2023diffuflow}, and FGTI~\cite{yangfrequency} leverage an external network to learn representations of condition inputs and use the cross-attention to capture deterministic information from conditional inputs. Further, to decrease the optimization difficulty in the denoising process, ControlTraj~\cite{zhu2024controltraj} pre-trains a road representation network with the masked modeling to embed topology constraints of road segments as conditional information for trajectory generation. USTD~\cite{hu2024towards} also pre-trains a spatio-temporal graph neural networks with the masked modeling to capture dependencies of conditions. Additionally, SaSDim~\cite{zhang2024sasdim} replaces the conventional stochastic differential equation with its probabilistic high-order form to more accurately estimate the variance of the noise in spatial time series and PhyDA~\cite{wang2025phyda} presents a physically regularized diffusion objective to keep consistency with physical laws in atmospheric systems, which can address the data generation bias. In the future, it is imperative to propose efficient denoising diffusion models that can deal with large-scale spatio-temporal data.

\section{Transferred Spatio-Temporal Foundation Models}\label{sec:transfer}
\emph{Transferred} spatio-temporal foundation models transfer the generalized knowledge of pre-trained foundation models trained on one or more other domains, such as language and vision, for spatio-temporal tasks. As shown in Fig.~\ref{pipeline_taxonomy}, to derive a transferred spatio-temporal foundation model, the pre-trained foundation model from other domains is first selected according to the corresponding data modalities. Then the pre-trained model is adapted to the specific spatio-temporal task and application by transforming the model or spatio-temporal data. In the following, we provide a detailed discussion about model selection and transfer adaption.

\subsection{Model Selection}
In this section, we divided the mainstream pre-trained foundation models into three categories according to the data modality, \emph{i.e.}, vision, language, and multi-modal models.
\subsubsection{Pre-Trained Vision Models}
There have been many vision foundation models in recent years that range from convolution-based ResNet~\cite{he2016deep} to Transformer-based ViT~\cite{alexey2020image}. They are trained on huge amounts of image data and thus can extract high-quality visual features of images for downstream tasks without training. Vision foundation models are popular as a backbone to preprocess frames of spatio-temporal video data. Early studies use pre-trained spatio-temporal convolution models to extract local visual and temporal features such as DCLR~\cite{ding2022dual}, which leverages the pre-trained R(2+1)D-18~\cite{tran2018closer} model to derive the general spatio-temporal representation of videos. Transformer architecture exihibits strong learning capabilities in computer vision. For example, MAESTL~\cite{feichtenhofer2022masked}, VideoMAE~\cite{tong2022videomae}, and CaPaint~\cite{duan2024causal} adopt the patch-based pre-trained vanilla vision Transformer to build a universal vision learner with the masked modeling. Additionally, LLM-AR~\cite{qu2024llms} and iVideoGPT~\cite{wu2024ivideogpt} firstly use a vector quantization technique (\emph{e.g.}, VQGAN and VQVAE) on spatio-temporal video data to acquire token sequences and then utilize tokens to train the Transformer-based foundation model. To further improve the video understanding capability, the large-scale pre-trained vision Transformer ViT-G~\cite{sun2023eva} is used in VideoChat~\cite{li2023videochat} and MovieChat~\cite{song2024moviechat}.

\subsubsection{Pre-Trained Language Models}
Pre-training a universal model to solve all language tasks has been studied for many years and led to a series of general methods. The most famous is ChatGPT developed by OpenAI~\cite{ray2023chatgpt}, which achieves powerful reasoning capabilities in various domains. Intuitively, the information of spatio-temporal data can be easily described with texts, for example, the time, location, event, and numerical value descriptions of these data. An intuitive way to leverage pre-trained language models is to utilize text descriptions of spatio-temporal data to chat with language models. After several interactions, the target spatio-temporal task will be addressed by leveraging the chain-of-thought composition capabilities of language models. Specifically, QWEN~\cite{bai2023qwen}, GLM~\cite{glm2024chatglm}, and GPT~\cite{ray2023chatgpt} are used in MAS4POI~\cite{wu2024mas4poi}, LLMob~\cite{wang2024large}, and ST-GIA+~\cite{zheng2024extracting} to understand human intentions in trajectories to predict future locations and generate realistic trajectories. Similarly, GPT, QWEN, and Mixtral~\cite{jiang2024mixtral} are used on spatio-temporal event data to understand causal relationships between events and perform meaningful predictions~\cite{yang2022large,yu2024onsep}. In addition, the general sequential modeling ability of pre-trained language models is considered in transferred foundation models, \emph{i.e.}, feeding the spatio-temporal data into language models. For imputation and prediction tasks, the masked and regressive pre-trained BERT~\cite{li2022spabert,wei2024ptr} and GPT-2~\cite{chen2023gatgpt,ren2024tpllm,liu2024can,wang2024empowering} are often utilized because of their interpolation and extrapolation capabilities. To take advantage of both reasoning and sequential modeling capabilities, AuxMobLCast~\cite{xue2022leveraging} uses BERT for learning context knowledge of historical trajectories and GPT-2 for prediction. Besides, Llama~\cite{touvron2023llama} is adopted in ST-LLM~\cite{liu2024spatial}, LLM4POI~\cite{li2024large}, and ExpTime~\cite{yuan2024back} to further improve the prediction accuracy.

\subsubsection{Pre-Trained Multi-Modal Models}
Spatio-temporal data often exist in multiple modalities, such as images and texts, in real-world scenarios. The pre-trained multi-modal model aims to align the representations of different modalities of data. In this way, images can be converted into texts and vice versa. To derive the text modality of satellite images as side information in urban region profiling, UrbanCLIP~\cite{yan2024urbanclip}, USPM~\cite{chen2024profiling}, and UrbanVLP~\cite{hao2024urbanvlp} utilize the pre-trained image-to-text model such as LLaMA-AdapterV2~\cite{gao2023llama} and SPHINX-V2~\cite{lin2023sphinx} to generate detailed descriptions of spatial information. Specifically, the image encoder of pre-trained CLIP~\cite{radford2021learning} is utilized on frames of video data~\cite{zhang2023video}, which enables language models recognizing visual representations for video understanding. This is because the image and text encoder of CLIP is aligned by contrastive learning. Similarly, the text encoder of CLIP is applied to the text condition for generating video~\cite{gu2024seer,chen2023videollm}. Moreover, image and text encoders are often collaboratively used to calculate CLIP scores, which is helpful for retrieving answers to video questions~\cite{pan2023retrieving} and navigating in street view~\cite{schumann2024velma}.

\subsection{Transfer Adaption}
Existing transferred adaption methods, as a form of transferring pre-trained foundation models for spatio-temporal tasks, employ various transformations on pre-trained models and primitive spatio-temporal data to bridge the gap between them, enabling effective transfer adaption. Despite significant differences in their design, transfer adaption in spatio-temporal tasks can be categorized into four representative approaches: \emph{prompt engineering}, \emph{feature enhancement}, \emph{cross-domain alignment}, and \emph{supervised fine-tuning}.
\begin{figure}[ht]
  \centering
\includegraphics[width=0.8\linewidth]{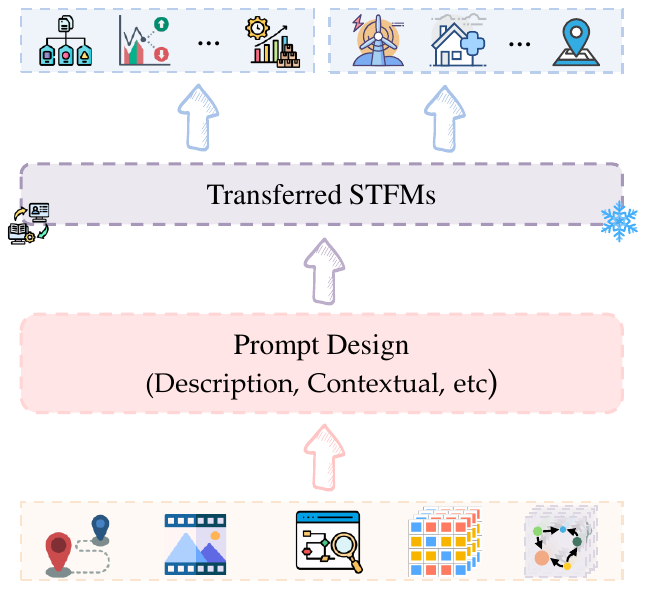}
  \caption{Prompt engineering, where \includegraphics[scale=0.07]{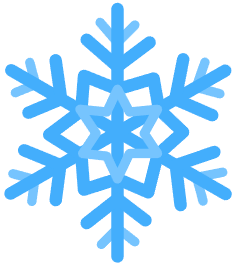} indicates frozen parameters.}
  \label{ta:pe}
\end{figure}
\subsubsection{Prompt Engineering}
The most important challenge for transferred spatio-temporal foundation models is to bridge the domain gap between original spatio-temporal data and pre-trained foundation models. It is intuitive to employ prompt engineering~\cite{giray2023prompt} to facilitate pre-trained foundation models understanding spatio-temporal data.

Existing prompt engineering methods for spatio-temporal foundation models can be divided into two categories: 1) directly incorporating numeric data into texts and 2) transforming numeric data into textual descriptions. The former directly embeds numeric values into predefined templates directly (\emph{e.g.}, "The average temperature is \{value\} degrees")~\cite{xue2023promptcast, xue2022leveraging, wang2023would}. In contrast, the latter converts numeric data into textual representations aiming to leverage the powerful memory capabilities of large language models. This is typically achieved by replacing numeric values with categorical labels based on predefined rules, such as using "D1" to denote data within the first bin~\cite{yu2023temporal}. These numeric-to-text transformations enable large language models to process quantitative data within their textual reasoning frameworks, albeit in a more coarse-grained manner. To further improve accuracy, recent studies emphasize enriching prompts with contextual information, such as historical trends, statistical summaries, or metadata relevant to the task~\cite{sun2023test, jin2023time, wang2023would, xue2022leveraging}. These methods are particularly valuable in domain-specific applications. For instance, in terms of stock price prediction, one study~\cite{yu2023temporal} constructs prompts, which include instructions, company profiles, historical temporal news summaries, and categorized stock price time series to enhance predictions. This design allows large language models to integrate cross-sequence information from multiple stocks and leverage their inherent knowledge to generate forecasts and explanations. In real-world applications, incorporating timely background information into contextual prompts helps the model identify temporal patterns and domain-specific nuances, leading to more accurate and robust performance. Additionally, few-shot examples are frequently employed to further enhance prediction performance~\cite{yu2023temporal, liu2023large}. For example, a limited number of samples (3-shot, 10-shot, and 25-shot) are provided to the large language model to contextualize it for the specific task~\cite{liu2023large}. These samples are embedded into textual templates to form question-answer pairs, which enhance the large language model in understanding the task structure and context. The large language model is subsequently trained on these few-shot examples to learn a task-specific prompt embedding, which is appended to each sample during inference. Moreover, TEST~\cite{sun2023test} uses soft prompting to refine prediction accuracy by generating task-specific embeddings that improve the large language model's comprehension of the input data. These soft prompts are trained using the loss derived from outputs of the large language model and ground truth of the task, effectively bridging the gap between the large language model and the specific requirements of time series forecasting tasks. According to existing studies~\cite{xue2023promptcast,yu2023temporal}, we observe that the performance of STFMs is highly dependent on manually designed prompts, and thus it requires improving the generalization of prompt engineering techniques in the future.

\begin{figure}[ht]
  \centering
\includegraphics[width=0.8\linewidth]{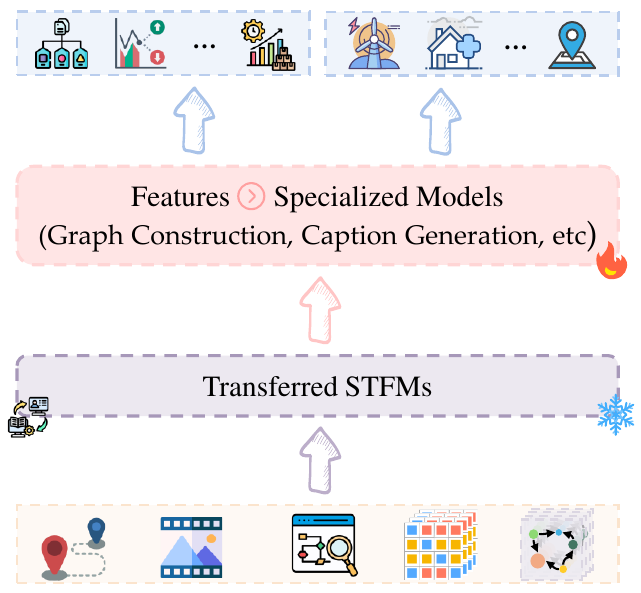}
  \caption{Feature enhancement, where \includegraphics[scale=0.07]{pics/snow.png} and \includegraphics[scale=0.07]{pics/fire.png} indicate frozen and active parameters, respectively.}
  \label{ta:fe}
\end{figure}
\subsubsection{Feature Enhancement}
Benefiting from the powerful text and image understanding capabilities of pre-trained foundation models, the high-level semantic features of spatio-temporal text and image data can be extracted for downstream tasks.

For example, VideoChat~\cite{li2023videochat} first extracts visual semantic features from video contents by the pre-trained video foundation model, and then feeds these features into large language models to conduct the video question answering. R2A~\cite{pan2023retrieving} further utilizes the pre-trained vision encoder of the CLIP model~\cite{radford2021learning} to extract semantic features of each frame of the input video, and then retrieve the top-$k$ similar texts to answer the question of the input video. In addition, UrbanCLIP~\cite{yan2024urbanclip} and USPM~\cite{chen2024profiling} use pre-trained image-to-text models~\cite{gao2023llama,li2024seed} to generate captions of urban images, which represents the text modality for the subsequent contrastive training. Moreover, CLIP-LSTM~\cite{wimmer2023leveraging} transforms the stock data into texts and images and then leverage the pre-trained CLIP encoders to extract the features from language and visual domains, which is beneficial to market forecasting. Further, ChatGNN~\cite{chen2023chatgpt} and RealTCD~\cite{li2024realtcd} directly leverage large language models to generate spatial correlations for spatio-temporal graph data. Specifically, ChatGNN feed the detail description of companies and daily media information into ChatGPT~\cite{ray2023chatgpt} to derive adjacency matrices of companies for stock prediction. RealTCD inputs the textual information of industrial systems into a large language model to reveal temporal causal relationships for anomaly detection. Different from ChatGNN and RealTCD, LA-GCN~\cite{xu2023language} utilizes the pre-trained BERT~\cite{kenton2019bert} to encode the texts of labels and joints of the skeleton data to build the distance-based adjacency matrix for skeleton-based action recognition. Next, PromptGAT~\cite{da2024prompt}, Orca~\cite{li2024ocean}, LLMPOI~\cite{liu2024semantic}, and ST-GIA+~\cite{zheng2024extracting} utilize a pre-trained large language model to generate predictions or embeddings as the input of the subsequent tasks based on the spatio-temporal prompts, \emph{e.g.}, weather and road information in spatio-temporal graph tasks, and locations and categories in spatio-temporal trajectory tasks. Finally, LLMob~\cite{wang2024large}, UrbanKGent~\cite{ning2024urbankgent}, and MAS4POI~\cite{wu2024mas4poi} use large language models to design a multi-agent collaboration system for spatio-temporal trajectory tasks. These methods feed the detailed description of trajectories into a large language model, which generates complete or fine-grained trajectories for recovery or up-sampling tasks, and then uses different large language models to refine or verify the previously generated trajectories. Although feature-enhanced STFMs achieved superior performance utilizing generated side information, it remains a problem to alleviate potential noise and fake information caused by the hallucination of pre-trained models.

\begin{figure}[ht]
  \centering
\includegraphics[width=0.8\linewidth]{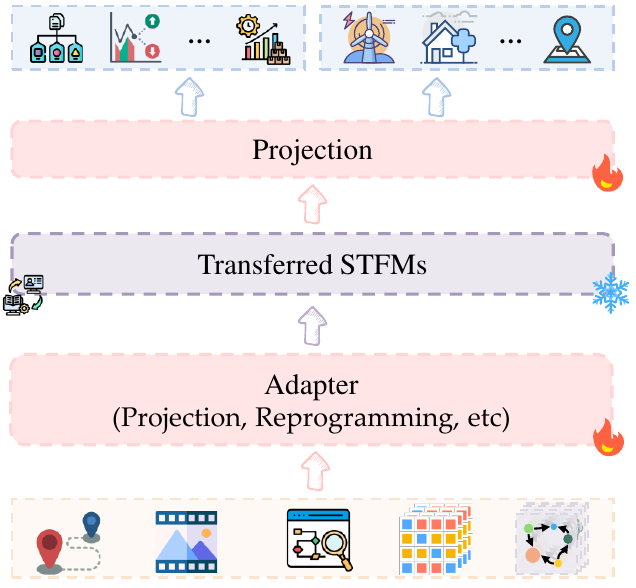}
  \caption{Cross-Domain alignment, where \includegraphics[scale=0.07]{pics/snow.png} and \includegraphics[scale=0.07]{pics/fire.png} indicate frozen and active parameters, respectively.}
  \label{ta:cda}
\end{figure}
\subsubsection{Cross-Domain Alignment}
Pre-trained large language models can also be directly used as the backbone for spatio-temporal modeling considering their powerful context inference capabilities. Existing methods often align the spatio-temporal data with the text and then feed the aligned data into frozen large language models for downstream spatio-temporal tasks.

VideoLLM~\cite{chen2023videollm} uses a trainable projection-based semantic translator to transform the visual spatial features into the textual embedding space of large language models for video understanding tasks. To inject temporal information into the aggregated frame-level representation for video understanding, MovieChat~\cite{song2024moviechat} and Video-LLaMA~\cite{zhang2023video} train a querying Transformer of the BLIP2~\cite{li2023blip} model before the projection layer. LLM-AR~\cite{qu2024llms} proposes a VQ-VAE-based~\cite{van2017neural} linguistic projection layer to reprogram the action signal in videos into language sentences through the human inductive bias, such as Zipf's law~\cite{piantadosi2014zipf}, and the hyperbolic codebook. VideoPoet~\cite{kondratyukvideopoet} utilizes the pre-trained text and visual foundation models to derive the aligned text and visual tokens. Additionally, FrozenBiLM~\cite{yang2022zero} adds a trainable projection layer between the frozen Transformer blocks of large language models, which enhances the video question answering due to the multi-cycle alignment. For the spatio-temporal graph tasks, embedding- and reprogramming-based alignment  in foundation models have been proposed. Specifically, STG-LLM~\cite{liu2024can} concatenated the time-of-day and day-of-week embeddings with the projected spatio-temporal graph enabling periodicity-aware large language models. REPST~\cite{wang2024empowering} aligns the spatio-temporal graph data with the textual embedding space by reprogramming the data from most correlated pre-trained word embeddings. TimeCMA~\cite{liu2025timecma} utilizes similarity retrieval-based cross-modality alignment to incorporate text information into time series. In addition, a trainable output projection layer~\cite{li2024urbangpt} is added into transferred foundation models to convert the transferred results into the original spatio-temporal domain. Although many existing methods focus on cross-domain data alignment, they mainly rely on simple addition and concatenation and fail to capture the complex relationships.

\begin{figure}[ht]
  \centering
\includegraphics[width=0.8\linewidth]{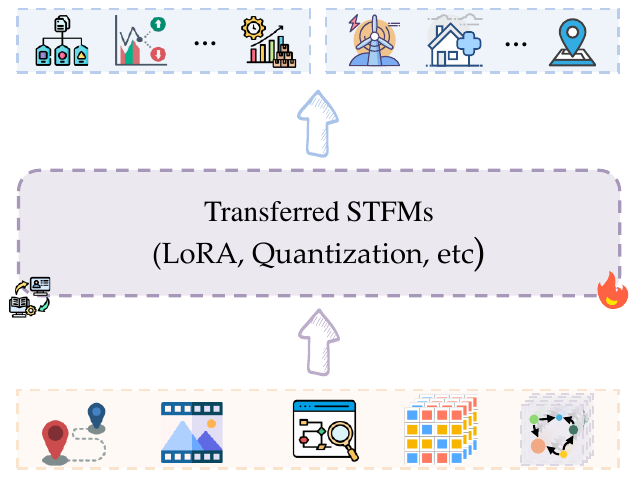}
  \caption{Supervised fine-tuning, where \includegraphics[scale=0.07]{pics/fire.png} indicates active parameters.}
  \label{ta:sft}
\end{figure}
\subsubsection{Supervised Fine-Tuning}
Supervised fine-tuning retrains the pre-trained model with the domain-specific data and has become a popular technique to align pre-trained foundation models with spatio-temporal data. This is because the data of the specific realm in the supervised fine-tuning is able to incorporate domain knowledge into the pre-trained large language models.

In the early stage, the supervised fine-tuning techniques have been adopted in the textual-related spatio-temporal tasks (\emph{e.g.}, geo-entity representation and geospatially language understanding). Specifically, SPABERT~\cite{li2022spabert} and GeoLM~\cite{li2023geolm} directly retrains full parameters of BERT~\cite{kenton2019bert} with geo-texts due to the same modality of the source model and target task. AuxMobLCast~\cite{xue2022leveraging} retrains all parameters in BERT-based encoder and GPT-based~\cite{ray2023chatgpt} decoder for human mobility forecasting, which aligns the source model and target task by transforming the numerical mobility data into language sentences. However, fine-tuning full parameters in pre-trained large language models is inefficient and be restricted in resource-constrained environments. To improve efficiency, partial fine-tuning methods emerge. Specifically, GATGPT~\cite{chen2023gatgpt} combines graph attention and a pre-trained large language model for spatio-temporal imputation, where the model only fine-tunes the weights of layer normalization modules in the pre-trained large language model during the supervised training stage. To preserve the sequential modeling capabilities acquired during pre-training , ST-LLM~\cite{liu2024spatial} further fine-tunes the last few attention layers in the pre-trained large language model for traffic forecasting. Moreover, TPLLM~\cite{ren2024tpllm} and PTR~\cite{wei2024ptr} incorporate the well-known parameter-efficient fine-tuning method, low-rank adaption~\cite{hulora}, into the pre-trained large language model for traffic forecasting and trajectory recovery, respectively. Specifically, in the supervised fine-tuning stage, two low-rank matrices are matrix multiplied to substitute original attention weights of the pre-trained large language model, which can significantly reduce the number of parameters and improve the efficiency of fine-tuning. To further decrease computation costs of fine-tuning, LLM4POI~\cite{li2024large} uses not only the low-rank adaption but also the 4-bit quantization technique for the next point-of-interest recommendation. While low-rank and quantization techniques reduce parameter redundancy during the supervised fine-tuning, they may also degrade model performance, particularly in complex tasks.

\section{Applications}\label{sec:app}
Next, we discuss primary application domains of STFMs. As shown in Fig.~\ref{param}, STFMs are widely adopted in various domains, encompassing energy, finance, weather, healthcare, transportation, and public services.

\subsection{Energy}
Primitive spatio-temporal foundation models for diverse energy applications have been designed including demand-side management, grid stability, and consumer behavior analysis. This is because of the abundant electricity time series and the inherent hierarchical spatial correlations of multiple electricity time series~\cite{tu2024powerpm}. Moreover, one of the most compelling use cases of STFMs in energy is in the predictive maintenance of renewable energy sources, like wind turbines and solar panels. For example, models trained on historical data of turbine performance and environmental conditions (such as wind speed and temperature) can predict when a turbine is likely to fail or require maintenance, which reduces downtime and improve the efficiency of wind farms by anticipating mechanical issues before they occur, optimizing maintenance schedules, and minimizing operational disruptions~\cite{fan2023spatio,ma2024fusionsf}.

\subsection{Finance}
Portfolio management~\cite{deng2024million}, fraud detection~\cite{duan2024cat}, and stock market prediction~\cite{daiya2024diffstock} are highly essential for the financial industry, and the implicit inter-relations across multiple time series of stock and supply-demand of enterprises are expected to be considered in them~\cite{hui2023constrained,luo2024timeseries}. Thus, proposed finance foundation models can capture not only temporal correlations but also spatial dependencies. In particular, spatial correlations is shown to provide better trading and investing~\cite{fang2024spatio}. Besides, it is worth noting that pre-trained language models can also assist financial analysis by deducing relationships between target entities~\cite{chen2023chatgpt}.

\subsection{Weather}
Spatio-temporal foundation models play an important role in weather-related applications. This is because a wide range of weather variables in the real-world such as precipitation vary with time and location, the seasonal and transmission properties of these weather variables are helpful for learning a spatio-temporal foundation model. To quickly and accurately solve global weather forecasting tasks at any time, spatio-temporal weather foundations models~\cite{bi2022pangu,chen2023fengwu} with a million parameters are pre-trained by massive European Centre for Medium-Range Weather Forecasts (ECMWF) reanalysis data. Moreover, spatio-temporal foundation models can be pre-trained by multi-modal weather data to further tackle various weather tasks such as super-resolution and image translation~\cite{zhao2024weathergfm}. These models have been instrumental in providing early warnings for disasters like Hurricane Dorian, which allowed for better-preparedness and evacuation efforts, saving lives and reducing property damage. In addition, STFMs have enhanced the ability to predict climate change impacts by analyzing long-term seasonal patterns, helping researchers and policymakers anticipate shifts in weather systems, which has been crucial for agricultural planning, water resource management, and disaster resilience. 

\subsection{Healthcare}
Electroencephalogram signals are typical spatio-temporal data due to neuronal transmissions and are pivotal for healthcare because many diseases can be detected through brain activity. To simultaneously address multiple tasks on electroencephalogram data with diverse formats, primitive foundation models with universal temporal represent abilities of electroencephalogram signals are pre-trained~\cite{yue2024eegpt}. Subsequently, primitive spatio-temporal foundation models are designed to further capture spatial dependencies between neural signals~\cite{zhang2024brant,jianglarge}. Moreover, electronic health records~\cite{xu2024protomix} are also key for healthcare AI systems by predicting clinical and operational events to assist decisions, and thus foundation models are pre-trained to solve all-purpose clinical tasks~\cite{yang2022large,jiang2023health,jiang2023think}.

\subsection{Transportation}
Massive amounts of transportation data have been generated by human, vehicles, and roads in recent years, which provides the possibility to use foundation models for various downstream transportation-related tasks to assist daily travel. Based on the spatial properties of transportation data reflected by connected road networks and the temporal properties of transportation data through mobilities, spatio-temporal modules should be introduced in transportation foundation models to improve performance. Specifically, STFMs have played a key role in intelligence transportation systems and autonomous vehicle systems. For intelligence transportation, STFMs have been used to analyze traffic patterns across time and space, which has allowed for the optimization of traffic lights and congestion managementl~\cite{lin2024trajfm,lai2025large}. This has led to smoother traffic flows, reduced travel times, and lower emissions. For autonomous vehicles, STFMs are crucial in enabling vehicles to understand both the spatial and temporal context of their environment—knowing not just where other cars are but also predicting their movements over time, improving both safety and efficiency in urban environments~\cite{yang2024generalized}.

\subsection{Public Services}
Public services are one of the greatest concerns to maintaining a harmonious society, textual and vision events that influence public services such as crimes and natural disasters are both closely related to geographic and seasonal variations. With the large amount of available event and video data, one-to-many primitive foundation models are pre-trained with spatio-temporal considerations for flood prediction~\cite{xu2024large}, earthquake monitoring~\cite{si2024all}, extreme weather forecasting~\cite{bi2022pangu}, and crime prediction~\cite{wu2024spatial}. To further leverage the multi-modal data, the transferred and primitive spatio-temporal foundation models are designed with pre-trained cross-domain foundation models~\cite{guo2024cross,si2024seisclip} and trained with multi-modal conditions~\cite{wang2024diffcrime} for seismic tasks and crime inference, respectively.

\section{Future Opportunities}\label{sec:oppo}
Although STFMs have achieved remarkable performance recently, several avenues for future work remain open.
\subsection{Scalability}
The scaling law in natural language processing inspired researchers to develop large language models with tens of billions, hundreds of billions, or even trillions of parameters. As for spatio-temporal foundation models, it is also vital to prove that performance is positively correlated with the data scale across different model sizes. The scalability of spatio-temporal foundation models determines whether this research line is meaningful (\emph{i.e.}, the large model represents the capability to improve performance as the data scale increasing) and provides valuable guidance for training and fine-tuning foundation models effectively. However, the size of current spatio-temporal foundation models is not sufficient to support the scaling law~\cite{li2024opencity,lin2024trajfm}. Larger models and more data are still needed to derive the emergence of LLM-like spatio-temporal capabilities.

\subsection{Efficiency}
Despite current achievements in spatio-temporal foundation models, they remain highly complex, leading to expensive computation needs. Particularly, the size of large-scale data-based primitive foundation models and large language models make them impractical to deploy for spatio-temporal applications in resource-constrained environments such as mobile phones or embedded systems. The trade-off between performance and efficiency of spatio-temporal foundation models is still a research problem, requiring foundation models to retain high one-to-many performance without too intricate designs and too large a size. Efficient deep learning methods such as distillation~\cite{zhang2024knowledge}, pruning~\cite{ma2023llm}, quantization~\cite{egashira2024exploiting}, condensation~\cite{miao2025less}, and deep learning units with subquadratic complexity~\cite{fang2024efficient} can be utilized in spatio-temporal foundation models in the future to make a balance between computation demands and one-to-many performance, which can bridge the gap between scientific research and industrial needs.

\subsection{Generalization}
Apart from using foundation models on the single spatio-temporal domain, extending foundation models for the spatio-temporal graph domain to other domains, such as the trajectory and spatio-temporal grid domains, provides considerable opportunities for future research. Most spatio-temporal grid data is generated by satellite images and spatio-temporal graph data is recorded by sensors. The difference between them is that sensors are irregularly distributed in the spatial dimension. Fortunately, universal spatial patch methods for the grid and graph data are proposed~\cite{fang2024efficient} and foundation models across these two domains are expected to be designed. Moreover, trajectories can be seen as univariate spatio-temporal graph data for integration into cross-domain foundation models~\cite{yu2024bigcity}, yet the irregular time interval and the spatial correlations of the same sequence should be carefully considered in future research. Addressing these issues will yield real universal spatio-temporal foundation models, which are more versatile and applicable to a broader range of tasks. Developing foundation models that effectively incorporate many spatio-temporal domains will be essential for emerging LLM-like spatio-temporal intelligence and further reducing energy consumption.

\subsection{Lack of Benchmarks}
Different from the diverse application-crossed tasks that spatio-temporal foundation models were applied for, dedicated benchmarks are provided for the time series forecasting task~\cite{qiu2024tfb}. Therefore, systematic evaluation and comprehensive benchmarks for all spatio-temporal domains need to be proposed urgently for a fair comparison of spatio-temporal foundation models. Proposing various and representative benchmarks that encompass a wide range of spatio-temporal domains, including event tasks, video tasks, trajectory tasks, spatio-temporal graph tasks, and spatio-temporal grid tasks, will offer a more thorough evaluation of the performance of STFMs. Systematic benchmarks will reflect the strengths and weaknesses of spatio-temporal foundation models in a fair comparison, promoting understanding and stimulating the improvement of spatio-temporal foundation models.

\subsection{Multi-Objective Training}
While using a single self-supervised training objective for primitive foundation models can achieve superior performance on simple spatio-temporal tasks, their performance on complex tasks is still limited, such as the task of predicting the future with missing historical data, which is difficult when using only regression or masked modeling. However, through combining denoising diffusion with masked modeling, spatio-temporal foundation models can obtain powerful generative capabilities while improving computational efficiency by masking. Moreover, incorporating different training objectives can capture complementary aspects of spatio-temporal data. For example, regression is good at learning effective temporal patterns and trends while masked modeling encourages contextual understanding. In addition, contrastive learning enhances feature discrimination and robustness, and denoising diffusion enhances generative capabilities. By combining these objectives, spatio-temporal foundation models are expected to gain stronger representation learning, better predictive capabilities, and improved robustness across different downstream tasks.

\subsection{Multi-Modal Foundation Models}
Currently, most spatio-temporal foundation models are learned from single modality data, \emph{e.g.}, time series, video, or grid. However, such single modality data might fall short in capturing important spatio-temporal context information and increases the difficulty in learning spatio-temporal general knowledge. Moreover, multi-modal knowledge from different sources can make models more robust to missing or noisy data. This may lead to better generalization, and enable various downstream tasks, such as autonomous driving. Besides, recent breakthroughs in foundation models (\emph{e.g.}, CLIP and VideoGPT) highlight the power of multi-modal integration. In fact, the multi-modal information can also be extracted from spatio-temporal data, \emph{e.g.}, heatmap images for trajectory data and text descriptions for event data. Therefore, multi-modal spatio-temporal foundation models are a feasible frontier research direction to improve robustness and generalization.

\section{Conclusion}\label{sec:conclu}
In this paper, we present a comprehensive survey of recent advancements in spatio-temporal foundation models, a growing research direction that solves various downstream spatio-temporal tasks via a single model. We offer an innovative pipeline perspective to categorize existing spatio-temporal foundation models. We first give an introduction to data harmonization for foundation models, providing researchers and practitioners with a holistic understanding to enhance the assessment of spatio-temporal data properties. Then primitive and transferred foundation models are introduced, covering the model designation and training objectives of primitive methods, and the model selection and transfer adaption of transferred methods. Finally, we underscore the restrictions of current spatio-temporal foundation models, while highlighting the opportunities in efficiency, scalability, generalization, and benchmarks. We expect that this survey not only explains the current state-of-the-art but also inspires further innovations and breakthroughs in the future.

\bibliographystyle{IEEEtran}
\bibliography{stfm}

\newpage

 




\vfill

\end{document}